\documentclass[journal]{IEEEtran}
\newcommand{\ignore}[1]{}
\usepackage[utf8]{inputenc}
\usepackage[T1]{fontenc}
\usepackage{amsmath,epsfig}
\usepackage{graphicx}
\usepackage{bm}
\usepackage{amssymb}
\usepackage{cite}
\usepackage{array}
\usepackage{extarrows}
\usepackage{url}

\usepackage{color}

\usepackage{multirow}
\usepackage{booktabs}
\usepackage{blkarray}
\usepackage{multirow}
\usepackage{array}
\usepackage{color}
\usepackage{pmat}

\begin{document}
%
\title{Deep Multiphase Level Set for Scene Parsing}
\author{Pingping~Zhang,
        Wei~Liu,
        Yinjie Lei,
        Hongyu Wang
        and~Huchuan~Lu
\thanks{
Copyright (c) 2019 IEEE. Personal use of this material is permitted. However, permission to use this material for any other purposes must  be obtained from the IEEE by sending an email to \textcolor{blue}{\underline{pubs-permissions@ieee.org}}.

PP. Zhang, HY. Wang and HC. Lu are with School of Information and Communication Engineering, Dalian University of Technology, Dalian, 116024, China.  (Email: jssxzhpp@mail.dlut.edu.cn; whyu@dlut.edu.cn; lhchuan@dlut.edu.cn) The corresponding author is Prof. Huchuan Lu.

W. Liu is with School of Computer Science, University of Adelaide, Adelaide, SA 50005, Australia. (Email: wei.liu02@adelaide.edu.au)

YJ. Lei is with College of Electronics and Information Engineering, Sichuan University, Chengdu, Sichuan,
610065, China. (Email: yinjie@scu.edu.cn)

This work is supported in part by the National Natural Science Foundation of China (NNSFC), No. 61725202, No. 61751212 and No. 61771088.
}
}
\maketitle
\markboth{IEEE Transactions on Image Processing}{}
\begin{abstract}
Recently, Fully Convolutional Network (FCN) seems to be the go-to architecture for image segmentation, including semantic scene parsing. However, it is difficult for a generic FCN to discriminate pixels around the object boundaries, thus FCN based methods may output parsing results with inaccurate boundaries. Meanwhile, level set based active contours are superior to the boundary estimation due to the sub-pixel accuracy that they achieve. However, they are quite sensitive to initial settings. To address these limitations, in this paper we propose a novel Deep Multiphase Level Set (DMLS) method for semantic scene parsing, which efficiently incorporates multiphase level sets into deep neural networks. The proposed method consists of three modules, i.e., recurrent FCNs, adaptive multiphase level set, and deeply supervised learning. More specifically, recurrent FCNs learn multi-level representations of input images with different contexts. Adaptive multiphase level set drives the discriminative contour for each semantic class, which makes use of the advantages of both global and local information. In each time-step of the recurrent FCNs, deeply supervised learning is incorporated for model training. Extensive experiments on three public benchmarks have shown that our proposed method achieves new state-of-the-art performances.
\end{abstract}
\IEEEpeerreviewmaketitle
\section{Introduction}
\IEEEPARstart{S}{emantic} scene parsing, also known as semantic image segmentation, plays an important role in various applications of computer vision and image processing, such as autonomous driving, robotic control and scene render.
%
%
It aims to segment and parse an image into different regions associated with predefined semantic categories, such as sky, road, person, and bed.
Traditional parsing approaches are mainly driven by low-level cues such as image intensity, color or texture.
However, due to the weakness of the handcrafted features to the illumination and noise, these methods maybe fail under complex scenes.
In recent years, the parsing results have been significantly improved thank to the deeply learned features and large-scale scene annotations~\cite{cordts2016cityscapes,neuhold2017mapillary,zhou2017scene}.
Meanwhile, Fully Convolutional Network (FCN)~\cite{long2015fully} seems to be the go-to architecture, which can adaptively learn the hierarchical features and directly predict the semantic category for each pixel.
However, it is difficult for a generic FCN to discriminate pixels around the object-oriented boundaries, thus FCN based methods may output segmentation maps with inaccurate boundaries.
To remedy this problem, many useful methods~\cite{yu2016multi,wang2016saliency,badrinarayanan2017segnet,zhao2017pyramid,chen2018deeplab,li2018referring,bilinski2018dense,zhang2018fully} have been proposed to integrate multi-scale or multi-level features for detail enhancements.
Although effective, current methods still lack of explicitly modeling complex topological boundaries.
Thus, there is still a large room for performance improvement.

Traditionally, active contour based methods are widely applied in image segmentation due to its admirable ability to handle object boundary changes~\cite{zhao1996variational,cremers2007review,olszewska2008multi,olszewska2015active}.
As a subclass of active contour based methods, Variational Level Set (VLS) based methods are superior to the boundary estimation due to the sub-pixel accuracy that they achieve.
Many remarkable works~\cite{mumford1989optimal,zhao1996variational,vese2002multiphase,cremers2007review,zhang2010variational} have shown the potential of VLS methods in achieving highly accurate image segmentation.
However, traditional VLS based methods are largely handicapped in capturing variations of real-world objects due to its sole dependency on gray values of images.
Besides, VLS based methods are hardly able to segment multiple objects with semantic information.
In order to overcome these limitations, multiple independent level set functions~\cite{vese2002multiphase,cremers2007review,zhang2010variational} are proposed to segment the specific region with each level set.
To make the segmentation compact, the formulation is modified so that a level set could not expand into a region already occupied by another.
However, if a part of the region is already occupied by a different level set, this
region will not be properly segmented.

To address above issues, in this work we propose a new formulation of VLS methods, called Deep Multiphase Level Set (DMLS), under the deep learning framework.
It combines the advantages of both deep learning and VLS methods for semantic scene parsing.
To achieve this goal, we first propose a novel Multiphase Level Set (MLS) method, which is a generalization of binary active contour models.
Different from previous methods, \emph{it can effectively handle the large-scale semantic categories and simultaneously avoid the presence of overlapped and void regions.}
To boost the representation abilities, then we incorporate the proposed MLS into deep leaning models with three key modules: Recurrent FCNs (RFCN), Adaptive MLS and Deeply Supervised Learning (DSL).
More specifically, the RFCN extends the vanilla FCN with recurrent connections to sequentially extract multi-level features, and treats coarse predictions as priors, which encode richer spatial and contextual information.
The adaptive MLS is formulated as a new layer that drives the contour for each predefined semantic categories, such that the energy function attains a minima for boundary prediction.
The DSL takes into account the importance of multiple level features for pixel classification, thus effectively merges coarse predictions.
Finally, with the help of all differentiable operations, an end-to-end network is built for robust scene labeling.
To the best of our knowledge, this is the first work that integrates MLS-based methods under a deep learning framework for scene parsing.

In summary, \textbf{our contributions} are three folds:
\begin{itemize}
\item
We propose a novel MLS framework for large-scale multi-class scene parsing.
The proposed framework can be seamlessly incorporated into any deep FCNs, providing robust semantic parsing.
\item
We propose a new multi-stage learning formulation for MLS with recurrent structures.
It turns the optimization process into an end-to-end learning.
When compared with previous level set methods, our formulation improves both segmentation effectiveness and accuracy.
\item
Extensive experiments on three public scene parsing benchmarks demonstrate that our approach achieves superior performance and performs better than very recent state-of-the-art methods.
\end{itemize}

The rest of this paper is organized as follows.
In Section II, we give an overview of classic level set and FCN based scene parsing methods.
Then we introduce the proposed approach in Section III.
In Section IV, we evaluate and analyze the proposed methods by extensive experiments and comparisons with other methods.
Finally, we provide the conclusion and future work in Section V.
\section{Related Work}
In this section, we briefly review scene parsing methods based on the classic level set and FCNs.
For more details, we refer the readers to the comprehensive survey~\cite{zhu2016beyond,huang2018learning}.
\vspace{-4mm}
\subsection{Classic Level Set Methods}
For scene parsing, the basic idea of level set based methods is to define an implicit function which represents object contours as the zero level set~\cite{sethian1999level,cremers2007review}.
The function is referred as the level set function, and is evolved according to a partial differential equation (PDE) derived from a Lagrangian formulation of active contour models.
However, early PDE driven level set methods utilize edge information, and are usually sensitive to the initialization and noises.
To solve this problem, variational methods~\cite{zhao1996variational,zhang2010variational} are proposed to derive the evolutional PDE directly from a certain energy function.
With an appropriate regularization, additional vision information like target region, or shape prior can be conveniently
and naturally formulated into the level set domain~\cite{cremers2007review}.
Mathematically, level set based segmentation problems can be solved by iteratively applying the gradient descent to minimize the evolution energy.
These properties make them suitable to be combined with deep networks
to solve the binary segmentation problem~\cite{hu2017deep}.
However, using this formulation for the contour limits the image partition to two regions (usually foreground and background).
Recently, there have been several approaches to segment more than two regions at the same time.
The most straightforward solution would be to employ multiple embedding functions, and then evolve
them independently.
However, they may fail when two regions overlap and/or voids appear.
In order to overcome this limitation, multiple level set functions~\cite{mumford1989optimal,zhao1996variational,vese2002multiphase,cremers2007review,zhang2010variational} are proposed to segment each class with an independent level set function.
However, in existing methods if a part of the region is already occupied by a different level set, this region will not be properly segmented.
As an interesting direction, parametric active contour based methods are also proposed.
For example, Olszewska \emph{et al.} propose the multi-feature vector flow~\cite{olszewska2008multi} and multi-layered representation~\cite{olszewska2011ontology,olszewska2013semantic} for scene understanding.
However, they are based on traditionally handcrafted features.

To address the aforementioned problems, in this work we present a novel look of level set methods under the view of deep learning frameworks.
The classic level set formulation is extended to the MLS framework with deeply learned features.
Thanks to the differentiable operations, our MLS can be easily combined with a fully end-to-end learning system, to efficiently handle the scene parsing task.
\vspace{-4mm}
\subsection{FCN based Methods}
Recently, with the advances of deep learning, various deep neural networks have been applied to semantic scene parsing and achieved
state-of-the-art performance.
For example, Long \emph{et al.}~\cite{long2015fully} first transform a classification network to output a spatial pixel-wise prediction, by replacing fully connected layers with convolutional layers.
Moreover, to improve the spatial details, they also fuse the coarse and high-level information to the fine and low-level information.
After that, many outstanding methods~\cite{badrinarayanan2017segnet,noh2015learning,zhou2017scene,zhao2017pyramid,chen2018deeplab,zhang2018context} are proposed to further incorporate multi-scale feature manipulation, dilation convolution or robust post-processing.
Among them, one important architecture for segmentation is based on the encoder-decoder structure.
For example, SegNet~\cite{badrinarayanan2017segnet} utilizes the max-pooling indices to perform non-linear upsampling, which eliminates the need for learning to upsample.
U-Net~\cite{ronneberger2015u} and SharpMask~\cite{pinheiro2016learning} transform the raw image into different level features through a Laplacian pyramid, and the feature maps of all scales are stage-wisely concatenated to form the robust representation for the final prediction.

More expressively, Zhao \emph{et al.}~\cite{zhao2017pyramid} and Chen \emph{et al.}~\cite{chen2018deeplab,chen2018encoder} develop the Atros Spatial Pyramid Pooling (ASPP) module, which can extract multi-scale features in a single network.
However, the dilated convolutions used in their works may result in local information missing and ``grids'' phenomena, which could affect the local consistency of feature maps~\cite{wang2018understanding}.
To enrich the context information, Zhang \emph{et al.}~\cite{yuan2018ocnet} introduce the context encoding module, which captures the semantic context of complex scenes and selectively highlights class-dependent feature maps.
Yuan \emph{et al.}~\cite{yuan2018ocnet} propose the pyramid object context to model the category dependencies.
Fu \emph{et al.}~\cite{fu2018moe} propose a mixture-of-experts based scene parsing network that incorporates a convolutional mixture-of-experts layer to assess the importance of features from different levels.
Zhao \emph{et al.}~\cite{zhao2018psanet} propose the point-wise spatial attention network to relax the local neighborhood constraint.
Furthermore, Zhang \emph{et al.}~\cite{zhang2019deep} propose a spatial gated attention module, which automatically highlights the attentive regions, resulting in accurate scene parsing.
Although effective, current methods still lack of explicitly modeling complex topological boundaries.
Compared to previous FCN based methods, our proposed framework inherits all the merits of the VLS model (to enrich the object curvatures) and the FCN based model (to encode powerful visual representation).
The proposed method is fully end-to-end trainable, and is able to learn both object contours via the MLS energy minimization and visual representation via the recurrent deep features.
\begin{figure*}
\centering
\resizebox{0.9\textwidth}{!}
{
\begin{tabular}{@{}c@{}c@{}}
\includegraphics[width=1\linewidth,height=10cm]{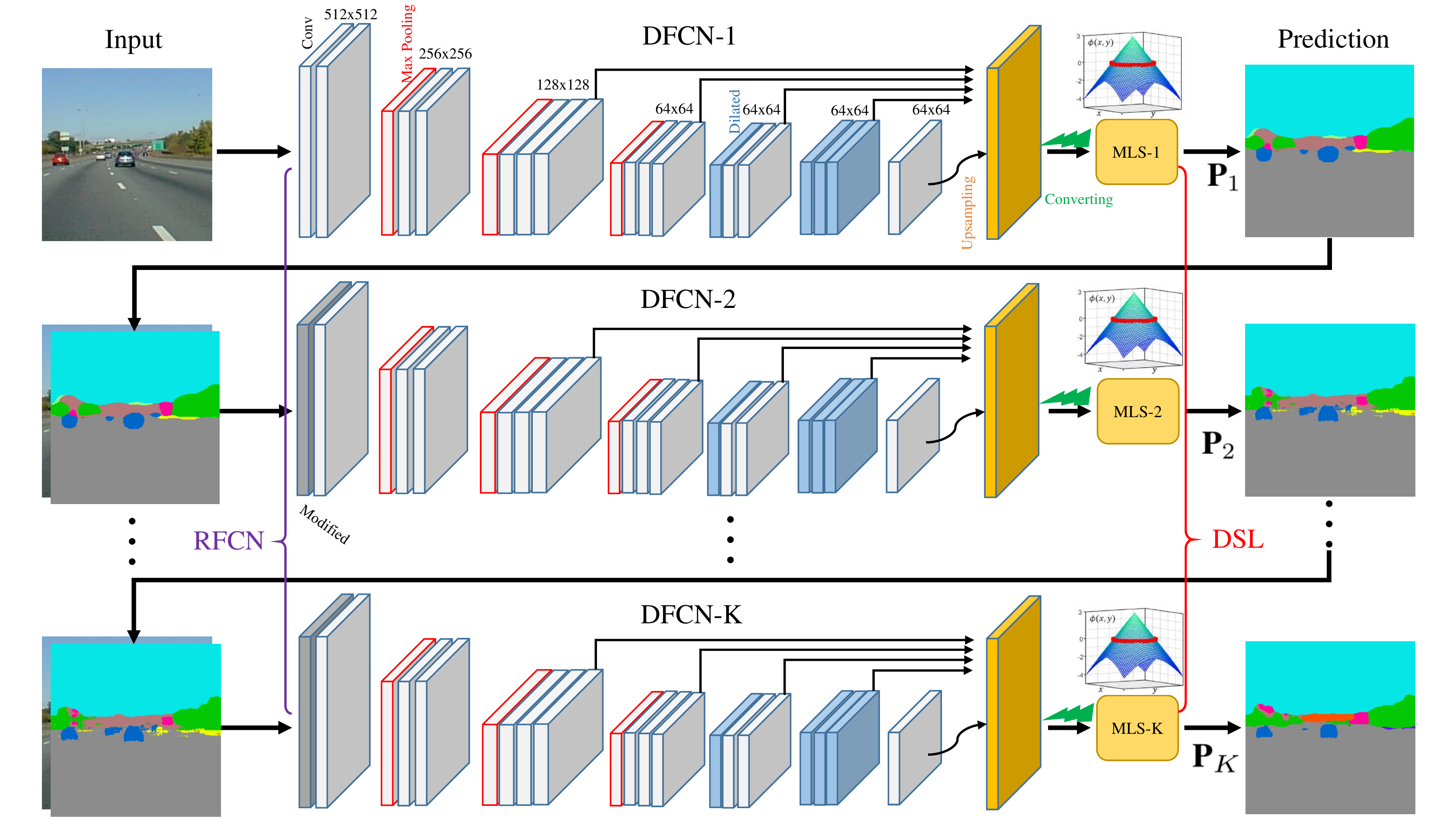} \\
\end{tabular}
}
\caption{The overall pipeline of our approach (DMLS). For this illustration, we adopt the VGG-16 backbone~\cite{simonyan2014very}. More details can be found in the main text.}
\vspace{-2mm}
\label{fig:framework}
\end{figure*}
\section{The Proposed Approach}
Fig.~\ref{fig:framework} illustrates the overall pipeline of our proposed approach (DMLS).
It consists of three key modules, \emph{i.e.}, Recurrent Fully Convolutional Network (RFCN),
Multiphase Level Set (MLS), and Deeply Supervised Learning (DSL).
Specifically, the RFCN includes multiple Dilated Fully Convolutional Networks (DFCN), which sequentially learn multi-level representations with spatial and contextual information.
The MLS captures spatial dependencies and boundaries at multiple levels, and drives the discriminative contour for each semantic class.
In each time-step of the RFCN, DSL is incorporated to effectively merge prediction maps, which take into account the importance of different levels for pixel classification.
With the help of all differentiable operations, an end-to-end network is built for scene labeling.
In this section, we first introduce the RFCN for deep sequential feature extraction.
Then, we provide the detailed formulation of the proposed MLS for multi-class segmentation.
Finally, we describe the DSL method for model training.
\subsection{Recurrent Fully Convolutional Network}
\begin{figure}
\centering
\resizebox{0.5\textwidth}{!}
{
\begin{tabular}{@{}c@{}c@{}}
\includegraphics[width=1\linewidth,height=4cm]{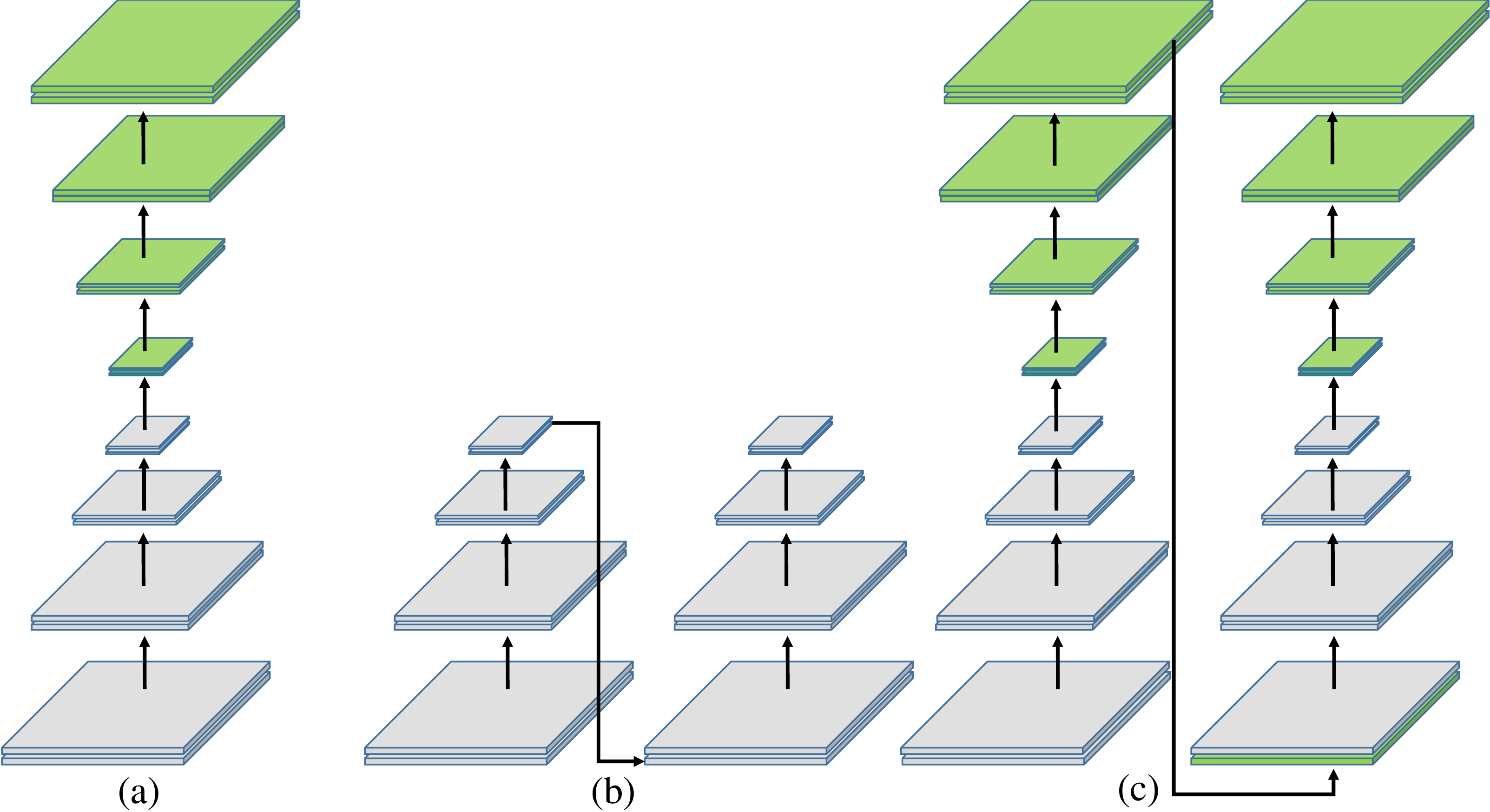} \\
\end{tabular}
}
\caption{Comparison of different deep models for scene parsing. (a) Encoder-decoder FCN. (b) Recurrent CNN~\cite{pinheiro2014recurrent}. (c) The proposed RFCN.}
\vspace{-4mm}
\label{fig:RFCN}
\end{figure}
Though expressive parsing performance has been achieved by various FCNs, there are still two major issues in previous methods~\cite{long2015fully,shen2017semantic,pohlen2017fullresolution}.
1) Most of existing FCNs predict the semantic label of a pixel only considering limited receptive fields.
They usually fail to enforce the spatial consistency, and may inevitably make incorrect predictions.
2) With only feed-forward architectures, FCNs can hardly refine the output predictions with accurate boundaries.

To mitigate the above issues, we propose the RFCN, which is an extension of vanilla FCN architectures.
The proposed RFCN extracts deep multi-level features sequentially, and enhance the context-detail information.
The comparison of different deep models is shown in Fig.~\ref{fig:RFCN}.
Most of existing methods follows the encoder-decoder FCN structure (Fig.~\ref{fig:RFCN} (a)).
Meanwhile, Pinheiro et al.~\cite{pinheiro2014recurrent} and Wang et al.~\cite{wang2016saliency} propose the Recurrent Convolutional Neural Network (RCNN), which also adopts the built-in recurrence, as shown in Fig.~\ref{fig:RFCN} (b).
However, there are key differences between the RCNN and our RFCN:
1) The RCNN aims to model the context information, which directly takes as input the high-level features with different resolutions.
While our RFCN incorporates previous predictions as spatial priors (see the right bottom of Fig.~\ref{fig:RFCN} (c)), which additionally captures the global-local information.
2) To build models, the RCNN is largely based on the complex CNNs.
While our RFCN is based on the cascaded fully convolutional architecture, which identifies and corrects its own errors.
3) Compared to the overloaded RCNN, we adopt the multi-level learning formulation.
With recurrent structure, our RFCN will slightly increase the training time.
However, due to the shared weights, our RFCN has fewer parameters, which make our proposed model easy for deployment.
%
%
It significantly improves both the segmentation effectiveness and accuracy than the RCNN.

In the first time step of the proposed RFCN, it only takes the raw image as input, and produces the coarse prediction.
In the following time steps, both the raw image and the coarse prediction are feed-forwarded to obtain the new prediction, which in turn serve as input in the next time step.
To incorporate more contexts and scene details, we consider the dilated convolutional architecture with skip connections, resulting to the Dilated Fully Convolutional Network (DFCN).
More details of the DFCN are shown in Fig.~\ref{fig:framework}.
Formally, each DFCN is divided into two parts, \emph{i.e.}, encoder part $E(\cdot,\theta_{e})$ and decoder part $D(\cdot,\theta_{d})$.
Our RFCN incorporates the previous prediction $P$ into the encoder part by modifying the first convolutional layer as
\begin{equation}
E(I,P)=W_{I}*I+W_{P}*P+b,
\label{equ:1}
\end{equation}
where $I$ and $P$ denote the input image and the previous prediction, respectively;
$W_{I}$ and $W_{P}$ represent corresponding convolution kernels; $b$ is the bias parameter.
In the first time step, $P_{1} = D(E(I; \theta_{e}); \theta_{d})$.
In the following time step, the RFCN refines the prediction by considering both the input image and the last coarse prediction as
\begin{equation}
P_{k+1} = D(E(I,P_{k}; \theta_{e}); \theta_{d}),k=1,...,K.
\label{equ:2}
\end{equation}
Our proposed RFCN architecture has three advantages over existing methods:
1) Coarse predictions are explicitly exploited to make training more easier, and yield large receptive fields for more accurate predictions;
2) In contrast to previous works, the output of our RFCN is reused as the feedback signal, such that the RFCN is capable to refine the prediction by correcting its previous mistakes until producing the final prediction in the last time step.
3) The proposed RFCN exploits both local features as well as more global contextual features simultaneously.
In our experiments, we observe that the segmentation accuracy almost converges after the fourth time step.
To balance the accuracy and efficiency, we set the total time step of the RFCN to $K =4$.
\subsection{Multiphase Level Set Formulation}
\begin{figure}
\centering
\resizebox{0.5\textwidth}{!}
{
\begin{tabular}{@{}c@{}c@{}}
\includegraphics[width=1\linewidth,height=3cm]{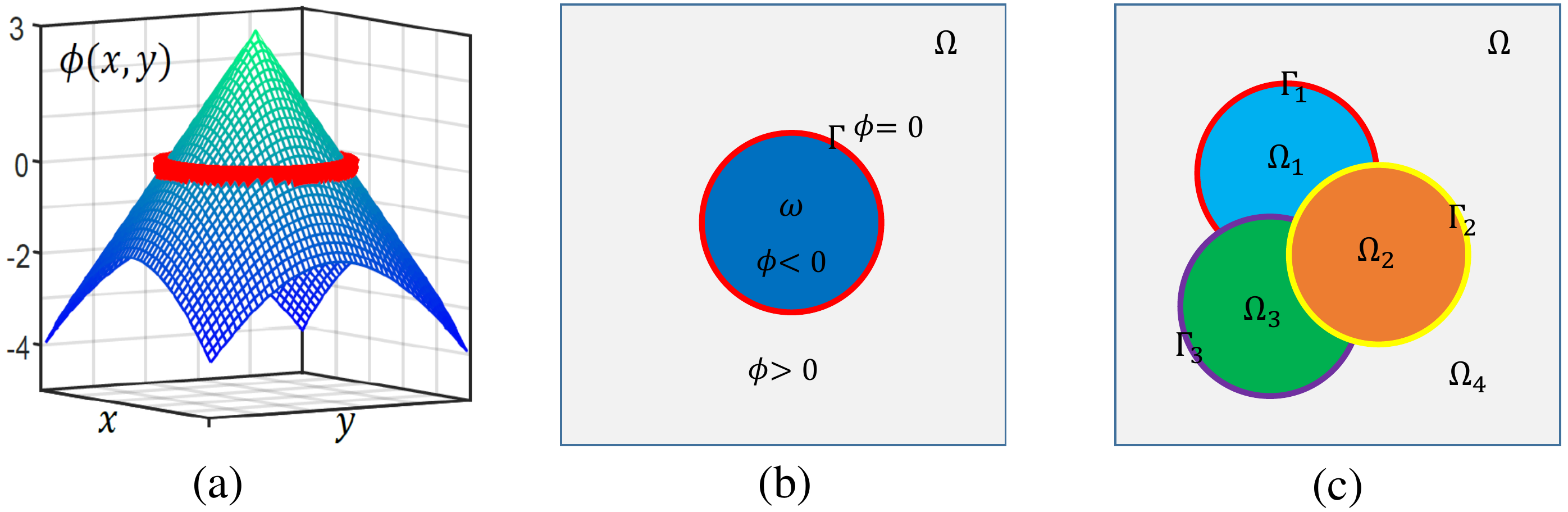} \\
\end{tabular}
}
\caption{Illustration of the level set methods. (a) Visualization of a level set function $\phi$ on the 2D image space. (b) The binary segmentation in the image domain $\Omega$. The zero level set and corresponding segmentation contour $\Gamma$ are marked in red. (c) Our MLS divides the $\Omega$ into multiple disjoint regions.}
\vspace{-4mm}
\label{fig:MLS}
\end{figure}

Traditional level set methods are superior to the boundary estimation due to the sub-pixel accuracy that they achieve.
However, they are quite sensitive to initial settings.
Besides, most of level set methods are used for binary segmentations.
Nevertheless, scene parsing needs pixel-wise labeling with abundant classes.
To overcome these limitations, we propose a novel MLS formulation, which can be incorporated into deep networks and deal with the multi-class scene parsing.
For training, the proposed MLS uses the gradient descent, which is seamless for deep learning framework.
Formally, let an image $I(\textbf{x})$ be defined on domain $\Omega\in\mathbb{R}^{M}$, and $\Omega_{i}, i=1,\cdots,N$ be the segmented regions.
We define a vector function $\Phi(\textbf{x}):\Omega\rightarrow\mathbb{R}^{N}$ as:
\begin{equation}
\textbf{x}=(x^{1},\cdots,x^{M})^{T}=(\phi^{1}(\textbf{x}),\cdots,\phi^{N}(\textbf{x}))^{T}.
\label{equ:3}
\end{equation}
If $\mathbb{R}^{N}$ is partitioned into $N$ regions, each of which has an assigned $\Omega_{i}$, then each point $\textbf{x}$ will belong to one and only one $\Omega_{i}$, according to the value of $\Phi(\textbf{x})$.
In the scalar valued case ($N=1$), the partition is assigning $\textbf{x}$ to specific disjoint regions by
\begin{equation}
  \label{equ:equ4}
\forall \textbf{x}\in \Omega
\left\{
\begin{aligned}
&\Gamma,&\phi(\textbf{x})=0\\
&inside(\Gamma),&\phi(\textbf{x})<0\\
&outside(\Gamma),&\phi(\textbf{x})>0\\
\end{aligned}
\right.
\end{equation}
where the interface $\Gamma$ is defined as the contour of an open subset $\omega$, and thus $\Gamma=\partial\omega$.
An example is shown in Fig.~\ref{fig:MLS} (b).
For the scalar valued function $\phi(\textbf{x})$, the classic level set evolution is given by a PDE whose expression is:
\begin{equation}
\phi_{t}+F(\textbf{x})|\nabla\phi|=0,\\
\phi(\textbf{x},t=0)=\phi_{0}(\textbf{x}),
\label{equ:5}
\end{equation}
where $F(\textbf{x})$ is a curvature-dependent function, $t$ is the evolution time,
and $\phi_{0}(\textbf{x})$ is the initial condition for $\phi_{\textbf{x},t=0}$.
For $N>1$, the level set method should divide the image domain $\Omega$ to more than three regions.
It is the requirement for the multi-class scene parsing.
To handle this task, we propose a new partition formulation, which avoids the overlap and/or void regions.
Formally, the partition is expressed as:
\begin{equation}
\label{equ:equ6}
\forall \textbf{x}\in \Omega
\left\{
\begin{aligned}
&\Gamma,&\phi^{i}(\textbf{x})=min_{j\neq i}(\phi^{j}(\textbf{x}))\\
&\Omega_{i},&\phi^{i}(\textbf{x})<min_{j\neq i}(\phi^{j}(\textbf{x})).
\end{aligned}
\right.
\end{equation}
We note that in this formulation the background is considered as an additional region.
Fig.~\ref{fig:MLS} (c) shows an example of our proposed partition with four divided regions.
We note that our partition is inspired by the reinitialization step in~\cite{vese2002multiphase}, where $\phi_{new}^{i}=\phi^{i}-max_{j\neq i}(\phi^{j}(\textbf{x}))$.
However, the method in~\cite{vese2002multiphase} can not well handle the void cases.
As shown in~\cite{vese2002multiphase}, it always provides wrong predictions when specific categories are not appeared in complex scenes.
In contrast, our method introduces an independent level set to each category, thus it can successfully handle the void cases and improve the segmentation accuracy.
Besides, our proposed MLS can be easily extended to large-scale semantic classes or disjoint regions.
Based on our proposed MLS, our method can parse complex images with more than 150 classes (see Section IV.D).
While previous methods only handles with few classes (<10), and usually process scene images with simple backgrounds.
As demonstrated in our experiments, our proposed MLS can be seamlessly incorporated into any deep FCNs for end-to-end learning.
While previous methods are based on handcrafted features, resulting in very low robustness.
Compared with classic level sets, our MLS is more robust and productive when dealing with the multi-class scene parsing.

In order to compute the evolution of the vector function $\Phi(\textbf{x})$, we propose a novel gradient learning formula for Eq.~5, which is applied to each component $\phi^{i}(\textbf{x})$.
Thus, our method can adaptively model the independent level sets, which is very different from previous works.
Formally, the corresponding PDE follows the expression:
\begin{equation}
\frac{\partial\phi^{i}}{\partial t}+(F^{i}(\textbf{x})-\varepsilon K^{i}(\Phi))|\nabla\phi^{i}|=0, i=1,\cdots,N,
\label{equ:7}
\end{equation}
where $F^{i}(\textbf{x})$ is the curvature propagation function for the $i$-th region, and $\varepsilon$ controls the weight of the regularization term $\textbf{K} =(K^{1},\cdots,K^{N})^{T}$.
Each $K^{i}(\Phi)$ is adapted for vector field regularization algorithms~\cite{lie2006binary}, and can be represented as:
\begin{equation}
K^{i}(\Phi)=\nabla((\frac{1}{N}\sum^{N}_{j=1}\frac{\nabla\phi^{j}}{|\nabla\phi^{j}|}\frac{(\nabla\phi^{j})^{T}}{|\nabla\phi^{j}|})\frac{\nabla\phi^{i}}{|\nabla\phi^{i}|}).
\label{equ:8}
\end{equation}
If we compare Equ.~\ref{equ:8} with the mean curvature of a scalar $\phi$, \emph{i.e.,} $\kappa(\phi)=\nabla(\frac{\nabla\phi^{i}}{|\nabla\phi^{i}|})$, we can see that
in our expression the normalized gradient is multiplied by
an $N\times N$ matrix$\frac{1}{N}\sum^{N}_{j=1}\frac{\nabla\phi^{j}}{|\nabla\phi^{j}|}\frac{(\nabla\phi^{j})^{T}}{|\nabla\phi^{j}|}$.
It ensures that the method counts every contour twice, and all the contours are treated equally.

The curvature propagation function $F^{i}(\textbf{x})$ is computed using a maximum likelihood approach.
To accomplish this goal, the probability density function (PDF) of $I(\textbf{x})$ for each region, $f_{I}(\textbf{x};\Theta^{i})$, is assumed to be a known priori.
Its parameters $\Theta^{i}$ are estimated from a number of previously selected points belonging to that region.
The expression of $F^{i}(\textbf{x})$ is
\begin{equation}
F^{i}(\textbf{x})=\log f_{I}(\textbf{x};\Theta^{i})-max(\{\log f_{I}(\textbf{x};\Theta^{i})\}_{\forall j\neq i}).
\label{equ:9}
\end{equation}
The purpose of this expression is to ensure that $F^{i}(\textbf{x})$ is positive at the locations where the pixel belongs to the $i$-th region, and negative otherwise.
The constraint is mainly based on the scene parsing task and FCN structure.
As shown in Section III.C, the proposed model predicts the confidence score, which measures how likely the pixel belong to a specific class.
Thus, the confidence score is nonnegative for all the classes.
To ensure the nonnegativity, we introduce the constraint by the max operation.
The expression can be modified to add a threshold $\rho$ under which the likelihood will be negative in any case:
\begin{equation}
F^{i}(\textbf{x})=\log f_{I}(\textbf{x};\Theta^{i})-max(\rho,\{\log f_{I}(\textbf{x};\Theta^{i})\}_{\forall j\neq i}).
\label{equ:9}
\end{equation}
This may be useful when the total number of regions present in the image is not a known priori.
In this case, an additional region would have to be created to contain those pixels whose PDFs for each region do not surpass the threshold.
\begin{table}[htbp]
\resizebox{0.42\textwidth}{!}
{
\begin{tabular}{p{0.65\linewidth}}
    \hline
    \textbf{Algorithm 1:} The Proposed DMLS Approach\\
    \hline
    \textbf{Input: } Raw image \textbf{I}.\\
    \textbf{Output: } Parsing result \textbf{P}.\\
    \hline
      \% Coarse segmentation by the first DFCN\\
      $\textbf{S} = DFCN(\textbf{I}; \theta_{e}); \theta_{d})$\\
      \% Convert predictions to [-0.5, 0.5]\\
      $\phi= \eta(\textbf{S})-\eta(1-\textbf{S})+\textbf{S}-0.5$\\
      \% \textbf{Recurrent Learning Procedure:}\\
      \textbf{for} $k=1,..,K$ \textbf{do} \\
      \quad $\phi_{old}=\phi$\\
      \quad \% MLS based curve evolution\\
      \quad $\phi_{new}=\phi_{old}+\frac{\partial\phi_{old}}{\partial t}$\\
      \quad \% Generate refined prediction\\
      \quad $\textbf{P}=assign(\phi_{new})$;\\
      \quad \% Coarse segmentation by the next DFCN\\
      \quad $\textbf{S} = DFCN(\textbf{I},\textbf{P}; \theta_{e}); \theta_{d})$\\
      \quad \% Convert predictions to [-0.5, 0.5]\\
      \quad $\phi= \eta(\textbf{S})-\eta(1-\textbf{S})+\textbf{S}-0.5$\\
      \textbf{end for}\\
     Return \textbf{P}\\
    \hline
\end{tabular}
}
\vspace{-4mm}
\label{alg:Alg1}
\end{table}

Because our proposed MLS is differentiable, thus it can be seamless to use the gradient descent for model training.
To enhance the representation abilities, we graft the proposed MLS onto each DFCN, which is more robust and powerful to deal with complex images.
More specifically, the coarse segmentation maps from DFCNs are converted to [-0.5, 0.5] via Euclidean distance transformation ($\eta$).
Then it is treated as a level set function $\phi= \eta(\textbf{S})-\eta(1-\textbf{S})+\textbf{S}-0.5$.
Based on the curve evolution of the level set function, we can recurrently refine the coarse predictions.
Algorithm 1 illustrates the overall recurrent learning procedure.
\subsection{Deeply Supervised Learning}
To train our model, we introduce the deeply supervised learning (DSL), which takes into account the importance of different levels for pixel classification.
As shown in Fig.~\ref{fig:framework}, we apply a weighted loss to each of the current steps.
The DSL can introduce multi-level supervision information to guide the MLS and DFCN modules.
Thus, it helps the proposed model to learn more discriminative features for scene parsing.
Formally, given the training dataset $D=\{(X_s,Y_s)\}^{S}_{s=1}$ with $S$ training image pairs, where $X_s =
\{x^s_i,i = 1,...,L\}$ and $Y_s = \{y^s_i,i = 1,...,L\}$ are the input image and the ground-truth segmentation with $L$ pixels, respectively.
$y^s_i = j(j\in N)$ denotes the labels of $j$-class.
For notional simplicity, we subsequently drop the subscript $s$ and consider each image independently.
In most of existing methods~\cite{long2015fully,zhao2017pyramid,chen2018deeplab}, the loss function used to train the network is the standard Softmax Cross-Entropy (SCE) loss.
Although effective, it mainly focuses on pixel-level accuracy and neglects the class distribution in images.
To automatically balance the loss between different semantic classes, we introduce weights $w_{j}$ (attributed to the $j$-class) to handle the class imbalance and the boundary errors.
Specifically, we define the following weighted cross-entropy (WCE) loss function,
\begin{equation}
\small{
\mathcal{L}_{wce}= -\frac{1}{L}\sum_{i=1}^{L}\sum_{j=0}^{N} In(y_{i}=j)w_{j}\text{log~Pr}(y_{i}=j|X;\theta)+\lambda||\theta||^{2}_2}.
\label{equ:11}
\end{equation}
where $\theta=(\theta_{e},\theta_{d}, \Theta)$ is the parameter of the network, $\lambda$ is the weight decay, $In(\cdot)$ is an indicator function and \text{Pr}$(y_i =j|X;\theta)=\frac{e^{y_i}}{\sum^C_{j=0}e^{y_j}}\in [0,1]$ is the confidence score of the network prediction that measures how likely the pixel belong to the $j$-class.
Given the frequency $f_j$ of class $j$ in the training data, the indicator function $In$, the training segmentation $\textbf{S}$, and the 2D gradient operator $\nabla$, the weights are defined as
\begin{equation}
\small{
w_{j}=\sum_{j}In(y_{i}=j))\frac{median(\textbf{f})}{f_j}+In(|\nabla \textbf{S}(i)|>0)},
\label{equ:12}
\end{equation}
where $\textbf{f} = [f_1,..., f_C]$ is the vector of all class frequencies.
The first term models median frequency balancing~\cite{badrinarayanan2017segnet}, and the second term assigns higher weights on the boundary regions.
The weighted loss function is continuously differentiable, so we can adopt the standard Stochastic Gradient Descent (SGD) method~\cite{krizhevsky2012imagenet} for model update.
Therefore, our approach is an end-to-end trainable architecture which does not need any post-processing procedures.
\section{Experiments}
Our proposed methods are evaluated on three public scene parsing benchmarks, including two traffic scene datasets (Cityscapes\footnote{\url{https://www.cityscapes-dataset.com}}~\cite{cordts2016cityscapes} and Mapillary Vistas\footnote{\url{https://research.mapillary.com}}~\cite{neuhold2017mapillary}) and one natural scene dataset (ADE20K\footnote{\url{http://groups.csail.mit.edu/vision/datasets/ADE20K}}~\cite{zhou2017scene}).
Following previous works~\cite{long2015fully,zhao2017pyramid,zhang2018context}, two widely-used metrics, \emph{i.e.}, pixel accuracy and mean Intersection over Union (IoU), are adopted to measure the parsing performance.
Let $n_{ij}$ be the number of pixels of class $i$ predicted to belong to class $j$, and $t_{i} =\sum_j n_{ij}$ be the total number of pixels of class $i$. $n_{cl}$ is the total classes in a dataset.
The two metrics are defined by:
\begin{itemize}
\item Pixel accuracy: $\sum_{i}n_{ii}/\sum_{i}t_{i}$.
\item Mean IoU: $(1/n_{cl})\sum_{i}n_{ii}/(t_{i}+\sum_{j}n_{ji}-n_{ii})$.
\end{itemize}
To fairly compare with other methods, we use either the public implementations with recommended parameter settings or the parsing results provided by the corresponding authors.
\subsection{Implementation Details}
Our approach is implemented with the public Caffe toolbox~\cite{jia2014caffe}.
The experiments are performed on a machine with four NVIDIA GTX TITAN X GPUs and an i4790 CPU.
Four DFCNs are utilized to build the RFCN.
For each DFCN, we follow the dilated convolutional architecture~\cite{yu2016multi}, which can be modified with different backbones.
For the dilated convolutions, we use $3\times3$ kernels with \emph{dilation=2}.
For training data, we uniformly resize each input image into 512$\times$512$\times$3 pixels.
Then, we randomly flip and crop the images to augment the training pairs.
The complete framework (including RFCN and MLS parts) is trained by the SGD~\cite{krizhevsky2012imagenet} with a momentum.
The weight decay $\lambda$ and momentum are set to 0.0005 and 0.9, respectively.
For the hyper-parameters of MLS, we experimentally find that using $\epsilon=1,\rho=0.5$ performs best.
The learning rate is initialized to be 10$^{-6}$, and decays exponentially with the rate of 0.1 after 20 epochs.
Meanwhile, higher learning rate (10$\times$) is used for newly-initialized parameters, \emph{i.e.}, MLS and recurrent layers.
The batch sizes for both training and testing phases are set to 1.
The results are reported after 40 training epochs.
It takes about three days to train our model.
When testing, our model runs at about 6.5 \emph{fps}.
%
We will release the source codes depending on the acceptance.
\begin{table}
\begin{center}
\doublerulesep=0.5pt
\caption{Quantitative results on the Cityscapes test set. The best two results are in bold and underline. Methods trained using both fine and coarse data are marked with $\ddagger$.}
\resizebox{0.48\textwidth}{!}
{
\begin{tabular}{c|c|c|c|c|c|c|c|c|c|c|c|c|c|c|c|c|c|c|c|c|c|c|c|c|c|c|c|c|||c|c|c|c|c|c|c|c|}
\hline
\multicolumn{4}{c|}{Methods} &\multicolumn{4}{|c|}{FCN backbone} &\multicolumn{2}{|c|}{IoU cla.\%}&\multicolumn{2}{|c|}{iIoU cla.\%}&\multicolumn{2}{|c|}{IoU cat\%}&\multicolumn{2}{|c}{iIoU cat.\%}
\\
\hline
\multicolumn{4}{c|}{FCN8s~\cite{long2015fully}}&\multicolumn{4}{|c|}{VGG-16}
&\multicolumn{2}{|c|}{65.3}&\multicolumn{2}{|c|}{41.7}&\multicolumn{2}{|c|}{85.7}&\multicolumn{2}{|c}{70.1}
\\
\multicolumn{4}{c|}{FCN8s~\cite{long2015fully}}&\multicolumn{4}{|c|}{ResNet-50}
&\multicolumn{2}{|c|}{69.4}&\multicolumn{2}{|c|}{41.6}&\multicolumn{2}{|c|}{86.3}&\multicolumn{2}{|c}{72.1}
\\
\multicolumn{4}{c|}{CRF-RNN~\cite{zheng2015conditional}}&\multicolumn{4}{|c|}{VGG-16}
&\multicolumn{2}{|c|}{62.5}&\multicolumn{2}{|c|}{34.4}&\multicolumn{2}{|c|}{82.7}&\multicolumn{2}{|c}{66.0}
\\
\multicolumn{4}{c|}{DilatedNet~\cite{yu2016multi}}&\multicolumn{4}{|c|}{VGG-16}
&\multicolumn{2}{|c|}{67.1}&\multicolumn{2}{|c|}{42.0}&\multicolumn{2}{|c|}{86.5}&\multicolumn{2}{|c}{71.1}
\\
\multicolumn{4}{c|}{LRR~\cite{ghiasi2016laplacian}}&\multicolumn{4}{|c|}{VGG-16}
&\multicolumn{2}{|c|}{69.7}&\multicolumn{2}{|c|}{48.0}&\multicolumn{2}{|c|}{88.2}&\multicolumn{2}{|c}{74.7}
\\
\multicolumn{4}{c|}{LRR$^{\ddagger}$~\cite{ghiasi2016laplacian}}&\multicolumn{4}{|c|}{VGG-16}
&\multicolumn{2}{|c|}{71.8}&\multicolumn{2}{|c|}{47.9}&\multicolumn{2}{|c|}{88.4}&\multicolumn{2}{|c}{73.9}
\\
\multicolumn{4}{c|}{CNN+CRF$^{\ddagger}$~\cite{lin2016efficient}}&\multicolumn{4}{|c|}{VGG-16}
&\multicolumn{2}{|c|}{71.6}&\multicolumn{2}{|c|}{51.7}&\multicolumn{2}{|c|}{87.3}&\multicolumn{2}{|c}{74.1}
\\
\multicolumn{4}{c|}{DSSPN$^{\ddagger}$~\cite{lin2016efficient}}&\multicolumn{4}{|c|}{ResNet-101}
&\multicolumn{2}{|c|}{76.6}&\multicolumn{2}{|c|}{56.2}&\multicolumn{2}{|c|}{89.6}&\multicolumn{2}{|c}{77.8}
\\
\multicolumn{4}{c|}{SegNet~\cite{badrinarayanan2017segnet}}&\multicolumn{4}{|c|}{VGG-16}
&\multicolumn{2}{|c|}{57.0}&\multicolumn{2}{|c|}{32.0}&\multicolumn{2}{|c|}{79.1}&\multicolumn{2}{|c}{61.9}
\\
\multicolumn{4}{c|}{FRRN~\cite{pohlen2017fullresolution}}&\multicolumn{4}{|c|}{ResNet-50}
&\multicolumn{2}{|c|}{71.8}&\multicolumn{2}{|c|}{45.5}&\multicolumn{2}{|c|}{88.9}&\multicolumn{2}{|c}{75.1}
\\
\multicolumn{4}{c|}{RefineNet$^{\ddagger}$~\cite{lin2017refinenet}}&\multicolumn{4}{|c|}{ResNet-152}
&\multicolumn{2}{|c|}{73.6}&\multicolumn{2}{|c|}{47.2}&\multicolumn{2}{|c|}{87.9}&\multicolumn{2}{|c}{70.6}
\\
\multicolumn{4}{c|}{PSPNet~\cite{zhao2017pyramid}}&\multicolumn{4}{|c|}{ResNet-101}
&\multicolumn{2}{|c|}{78.4}&\multicolumn{2}{|c|}{56.7}&\multicolumn{2}{|c|}{90.6}&\multicolumn{2}{|c}{78.6}
\\
\multicolumn{4}{c|}{PSPNet$^{\ddagger}$~\cite{zhao2017pyramid}}&\multicolumn{4}{|c|}{ResNet-101}
&\multicolumn{2}{|c|}{80.2}&\multicolumn{2}{|c|}{58.1}&\multicolumn{2}{|c|}{90.6}&\multicolumn{2}{|c}{78.2}
\\
\multicolumn{4}{c|}{SegModel$^{\ddagger}$~\cite{shen2017semantic}}&\multicolumn{4}{|c|}{ResNet-101}
&\multicolumn{2}{|c|}{79.2}&\multicolumn{2}{|c|}{56.4}&\multicolumn{2}{|c|}{90.4}&\multicolumn{2}{|c}{77.0}
\\
\multicolumn{4}{c|}{PEARL$^{\ddagger}$~\cite{jin2017video}}&\multicolumn{4}{|c|}{ResNet-101}
&\multicolumn{2}{|c|}{75.4}&\multicolumn{2}{|c|}{51.6}&\multicolumn{2}{|c|}{89.2}&\multicolumn{2}{|c}{75.1}
\\
\multicolumn{4}{c|}{SAC$^{\ddagger}$~\cite{zhang2017scale}}&\multicolumn{4}{|c|}{ResNet-101}
&\multicolumn{2}{|c|}{78.1}&\multicolumn{2}{|c|}{55.2}&\multicolumn{2}{|c|}{90.6}&\multicolumn{2}{|c}{78.3}
\\
\multicolumn{4}{c|}{DeepLabv2$^{\ddagger}$~\cite{chen2018deeplab}}&\multicolumn{4}{|c|}{ResNet-101}
&\multicolumn{2}{|c|}{70.4}&\multicolumn{2}{|c|}{42.6}&\multicolumn{2}{|c|}{86.4}&\multicolumn{2}{|c}{67.7}
\\
\multicolumn{4}{c|}{DUC~\cite{wang2018understanding}}&\multicolumn{4}{|c|}{ResNet-50}
&\multicolumn{2}{|c|}{77.6}&\multicolumn{2}{|c|}{53.6}&\multicolumn{2}{|c|}{90.1}&\multicolumn{2}{|c}{75.2}
\\
\multicolumn{4}{c|}{DRN~\cite{zhuang2018dense}}&\multicolumn{4}{|c|}{ResNet-101}
&\multicolumn{2}{|c|}{79.9}&\multicolumn{2}{|c|}{56.1}&\multicolumn{2}{|c|}{91.1}&\multicolumn{2}{|c}{79.4}
\\
\multicolumn{4}{c|}{DRN$^{\ddagger}$~\cite{zhuang2018dense}}&\multicolumn{4}{|c|}{ResNet-101}
&\multicolumn{2}{|c|}{82.8}&\multicolumn{2}{|c|}{61.1}&\multicolumn{2}{|c|}{91.8}&\multicolumn{2}{|c}{80.7}
\\
\multicolumn{4}{c|}{Inplace-ABN$^{\ddagger}$~\cite{rota2018place}}&\multicolumn{4}{|c|}{ResNet-101}
&\multicolumn{2}{|c|}{82.0}&\multicolumn{2}{|c|}{65.9}&\multicolumn{2}{|c|}{91.2}&\multicolumn{2}{|c}{81.7}
\\
\multicolumn{4}{c|}{DenseASPP$^{\ddagger}$~\cite{yang2018denseaspp}}&\multicolumn{4}{|c|}{DenseNet161}
&\multicolumn{2}{|c|}{80.6}&\multicolumn{2}{|c|}{57.9}&\multicolumn{2}{|c|}{90.7}&\multicolumn{2}{|c}{78.1}
\\
\multicolumn{4}{c|}{DPC$^{\ddagger}$~\cite{chen2018searching}}&\multicolumn{4}{|c|}{ResNet-101}
&\multicolumn{2}{|c|}{82.7}&\multicolumn{2}{|c|}{63.3}&\multicolumn{2}{|c|}{92.0}&\multicolumn{2}{|c}{82.5}
\\
\multicolumn{4}{c|}{DeepLabv3+$^{\ddagger}$~\cite{chen2018encoder}}&\multicolumn{4}{|c|}{ResNet-101}
&\multicolumn{2}{|c|}{82.1}&\multicolumn{2}{|c|}{62.4}&\multicolumn{2}{|c|}{92.0}&\multicolumn{2}{|c}{81.9}
\\
\multicolumn{4}{c|}{GANet~\cite{zhang2019deep}}&\multicolumn{4}{|c|}{ResNet-101}
&\multicolumn{2}{|c|}{81.6}&\multicolumn{2}{|c|}{63.6}&\multicolumn{2}{|c|}{91.6}&\multicolumn{2}{|c}{81.0}
\\
\multicolumn{4}{c|}{GANet$^{\ddagger}$~\cite{zhang2019deep}}&\multicolumn{4}{|c|}{ResNet-101}
&\multicolumn{2}{|c|}{\underline{82.8}}&\multicolumn{2}{|c|}{\textbf{66.4}}&\multicolumn{2}{|c|}{\underline{92.2}}&\multicolumn{2}{|c}{\underline{82.7}}
\\
\multicolumn{4}{c|}{WResNet~\cite{wu2016wider}}&\multicolumn{4}{|c|}{WResNet-38}
&\multicolumn{2}{|c|}{78.4}&\multicolumn{2}{|c|}{59.1}&\multicolumn{2}{|c|}{90.9}&\multicolumn{2}{|c}{81.1}
\\
\hline
\multicolumn{4}{c|}{DeepLabv3$^{\ddagger}$~\cite{chen2017rethinking}}&\multicolumn{4}{|c|}{ResNet-101}
&\multicolumn{2}{|c|}{81.3}&\multicolumn{2}{|c|}{62.1}&\multicolumn{2}{|c|}{91.6}&\multicolumn{2}{|c}{81.7}
\\
\multicolumn{4}{c|}{OCNet~\cite{yuan2018ocnet}}&\multicolumn{4}{|c|}{ResNet-101}
&\multicolumn{2}{|c|}{81.2}&\multicolumn{2}{|c|}{61.3}&\multicolumn{2}{|c|}{91.6}&\multicolumn{2}{|c}{81.1}
\\
\multicolumn{4}{c|}{Fast-OCNet~\cite{yuan2018ocnet}}&\multicolumn{4}{|c|}{ResNet-101}
&\multicolumn{2}{|c|}{82.1}&\multicolumn{2}{|c|}{61.0}&\multicolumn{2}{|c|}{91.7}&\multicolumn{2}{|c}{80.7}
\\
\hline
\multicolumn{4}{c|}{DMLS (Ours)}&\multicolumn{4}{|c|}{VGG-16}
&\multicolumn{2}{|c|}{72.8}&\multicolumn{2}{|c|}{49.7}&\multicolumn{2}{|c|}{89.4}&\multicolumn{2}{|c}{76.6}
\\
\multicolumn{4}{c|}{DMLS (Ours)}&\multicolumn{4}{|c|}{ResNet-50}
&\multicolumn{2}{|c|}{79.7}&\multicolumn{2}{|c|}{60.2}&\multicolumn{2}{|c|}{91.1}&\multicolumn{2}{|c}{79.5}
\\
\multicolumn{4}{c|}{DMLS$^{\ddagger}$ (Ours)}&\multicolumn{4}{|c|}{ResNet-50}
&\multicolumn{2}{|c|}{82.0}&\multicolumn{2}{|c|}{61.8}&\multicolumn{2}{|c|}{92.0}&\multicolumn{2}{|c}{81.7}
\\
\multicolumn{4}{c|}{DMLS (Ours)}&\multicolumn{4}{|c|}{ResNet-101}
&\multicolumn{2}{|c|}{83.1}&\multicolumn{2}{|c|}{64.6}&\multicolumn{2}{|c|}{92.4}&\multicolumn{2}{|c}{82.3}
\\
\multicolumn{4}{c|}{DMLS$^{\ddagger}$ (Ours)}&\multicolumn{4}{|c|}{ResNet-101}
&\multicolumn{2}{|c|}{\textbf{83.7}}&\multicolumn{2}{|c|}{\underline{66.1}}&\multicolumn{2}{|c|}{\textbf{93.0}}&\multicolumn{2}{|c}{\textbf{83.3}}
\\
\hline
\end{tabular}
}
\label{table:cityscapes_test}
\vspace{-4mm}
\end{center}
\end{table}
\begin{table*}
\begin{center}
\doublerulesep=0.5pt
\caption{Category-wise comparison on the Cityscapes test set. The best results are in bold. Methods trained using both fine and coarse data are marked with $\ddagger$.}
\resizebox{0.96\textwidth}{!}
{
\begin{tabular}{|c|c|c|c|c|c|c|c|c|c|c|c|c|c|c|c|c|c|c|c|c|c|c|c|c|c|c|c|c|c|c|c|c|c|c|c|c|c|c|c|c|c|c|c|c|c|c|c|c|c|c|c|c|c|c|c|c|c|c|c|c|c|c|c|c|}
\hline
\multicolumn{4}{|c|}{Methods}&\multicolumn{4}{|c|}{Mean IoU} &\multicolumn{2}{|c|}{\rotatebox{90}{road}}&\multicolumn{2}{|c|}{\rotatebox{90}{sidewalk}}
&\multicolumn{2}{|c|}{\rotatebox{90}{building}}&\multicolumn{2}{|c|}{\rotatebox{90}{wall}}
&\multicolumn{2}{|c|}{\rotatebox{90}{fence}}&\multicolumn{2}{|c|}{\rotatebox{90}{pole}}
&\multicolumn{2}{|c|}{\rotatebox{90}{traffic light}}&\multicolumn{2}{|c|}{\rotatebox{90}{traffic sign}}
&\multicolumn{2}{|c|}{\rotatebox{90}{vegetation}}&\multicolumn{2}{|c|}{\rotatebox{90}{terrain}}
&\multicolumn{2}{|c|}{\rotatebox{90}{sky}}&\multicolumn{2}{|c|}{\rotatebox{90}{person}}
&\multicolumn{2}{|c|}{\rotatebox{90}{rider}}&\multicolumn{2}{|c|}{\rotatebox{90}{car}}
&\multicolumn{2}{|c|}{\rotatebox{90}{truck}}&\multicolumn{2}{|c|}{\rotatebox{90}{bus}}
&\multicolumn{2}{|c|}{\rotatebox{90}{train}}&\multicolumn{2}{|c|}{\rotatebox{90}{motorcycle}}
&\multicolumn{2}{|c|}{\rotatebox{90}{bicycle}}
\\
\hline
\multicolumn{4}{|c|}{CRF-RNN~\cite{zheng2015conditional}}
&\multicolumn{4}{|c|}{62.5}&\multicolumn{2}{|c|}{96.3}
&\multicolumn{2}{|c|}{73.9}&\multicolumn{2}{|c|}{88.2}
&\multicolumn{2}{|c|}{47.6}&\multicolumn{2}{|c|}{41.3}
&\multicolumn{2}{|c|}{35.2}&\multicolumn{2}{|c|}{49.5}
&\multicolumn{2}{|c|}{59.7}&\multicolumn{2}{|c|}{90.6}
&\multicolumn{2}{|c|}{66.1}&\multicolumn{2}{|c|}{93.5}
&\multicolumn{2}{|c|}{70.4}&\multicolumn{2}{|c|}{34.7}
&\multicolumn{2}{|c|}{90.1}&\multicolumn{2}{|c|}{39.2}
&\multicolumn{2}{|c|}{57.5}&\multicolumn{2}{|c|}{55.4}
&\multicolumn{2}{|c|}{43.9}&\multicolumn{2}{|c|}{54.6}
\\
\multicolumn{4}{|c|}{FCN8s~\cite{long2015fully}}
&\multicolumn{4}{|c|}{65.3}&\multicolumn{2}{|c|}{97.4}
&\multicolumn{2}{|c|}{78.4}&\multicolumn{2}{|c|}{89.2}
&\multicolumn{2}{|c|}{34.9}&\multicolumn{2}{|c|}{44.2}
&\multicolumn{2}{|c|}{47.4}&\multicolumn{2}{|c|}{60.1}
&\multicolumn{2}{|c|}{65.0}&\multicolumn{2}{|c|}{91.4}
&\multicolumn{2}{|c|}{69.3}&\multicolumn{2}{|c|}{93.9}
&\multicolumn{2}{|c|}{77.1}&\multicolumn{2}{|c|}{51.4}
&\multicolumn{2}{|c|}{92.6}&\multicolumn{2}{|c|}{35.3}
&\multicolumn{2}{|c|}{48.6}&\multicolumn{2}{|c|}{46.5}
&\multicolumn{2}{|c|}{51.6}&\multicolumn{2}{|c|}{66.8}
\\
\multicolumn{4}{|c|}{DilatedNet~\cite{yu2016multi}}
&\multicolumn{4}{|c|}{67.1}&\multicolumn{2}{|c|}{97.6}
&\multicolumn{2}{|c|}{79.2}&\multicolumn{2}{|c|}{89.9}
&\multicolumn{2}{|c|}{37.3}&\multicolumn{2}{|c|}{47.6}
&\multicolumn{2}{|c|}{53.2}&\multicolumn{2}{|c|}{58.6}
&\multicolumn{2}{|c|}{65.2}&\multicolumn{2}{|c|}{91.8}
&\multicolumn{2}{|c|}{69.4}&\multicolumn{2}{|c|}{93.7}
&\multicolumn{2}{|c|}{78.9}&\multicolumn{2}{|c|}{55.0}
&\multicolumn{2}{|c|}{93.3}&\multicolumn{2}{|c|}{45.5}
&\multicolumn{2}{|c|}{53.4}&\multicolumn{2}{|c|}{47.7}
&\multicolumn{2}{|c|}{52.2}&\multicolumn{2}{|c|}{66.0}
\\
\multicolumn{4}{|c|}{LRR~\cite{ghiasi2016laplacian}}
&\multicolumn{4}{|c|}{69.7}&\multicolumn{2}{|c|}{97.7}
&\multicolumn{2}{|c|}{79.9}&\multicolumn{2}{|c|}{90.7}
&\multicolumn{2}{|c|}{44.4}&\multicolumn{2}{|c|}{48.6}
&\multicolumn{2}{|c|}{58.6}&\multicolumn{2}{|c|}{68.2}
&\multicolumn{2}{|c|}{72.0}&\multicolumn{2}{|c|}{92.5}
&\multicolumn{2}{|c|}{69.3}&\multicolumn{2}{|c|}{94.7}
&\multicolumn{2}{|c|}{81.6}&\multicolumn{2}{|c|}{60.0}
&\multicolumn{2}{|c|}{94.0}&\multicolumn{2}{|c|}{43.6}
&\multicolumn{2}{|c|}{56.8}&\multicolumn{2}{|c|}{47.2}
&\multicolumn{2}{|c|}{54.8}&\multicolumn{2}{|c|}{69.7}
\\
\multicolumn{4}{|c|}{LRR$\ddagger$~\cite{ghiasi2016laplacian}}
&\multicolumn{4}{|c|}{71.8}&\multicolumn{2}{|c|}{97.9}
&\multicolumn{2}{|c|}{81.5}&\multicolumn{2}{|c|}{91.4}
&\multicolumn{2}{|c|}{50.5}&\multicolumn{2}{|c|}{52.7}
&\multicolumn{2}{|c|}{59.4}&\multicolumn{2}{|c|}{66.8}
&\multicolumn{2}{|c|}{72.7}&\multicolumn{2}{|c|}{92.5}
&\multicolumn{2}{|c|}{70.1}&\multicolumn{2}{|c|}{95.0}
&\multicolumn{2}{|c|}{81.3}&\multicolumn{2}{|c|}{60.1}
&\multicolumn{2}{|c|}{94.3}&\multicolumn{2}{|c|}{51.2}
&\multicolumn{2}{|c|}{67.7}&\multicolumn{2}{|c|}{54.6}
&\multicolumn{2}{|c|}{55.6}&\multicolumn{2}{|c|}{69.6}
\\
\multicolumn{4}{|c|}{DeepLabv2+CRF~\cite{chen2018deeplab}}
&\multicolumn{4}{|c|}{70.4}&\multicolumn{2}{|c|}{97.9}
&\multicolumn{2}{|c|}{81.3}&\multicolumn{2}{|c|}{90.3}
&\multicolumn{2}{|c|}{48.8}&\multicolumn{2}{|c|}{47.4}
&\multicolumn{2}{|c|}{49.6}&\multicolumn{2}{|c|}{57.9}
&\multicolumn{2}{|c|}{67.3}&\multicolumn{2}{|c|}{91.9}
&\multicolumn{2}{|c|}{69.4}&\multicolumn{2}{|c|}{94.2}
&\multicolumn{2}{|c|}{79.8}&\multicolumn{2}{|c|}{59.8}
&\multicolumn{2}{|c|}{93.7}&\multicolumn{2}{|c|}{56.5}
&\multicolumn{2}{|c|}{67.5}&\multicolumn{2}{|c|}{57.5}
&\multicolumn{2}{|c|}{57.7}&\multicolumn{2}{|c|}{68.8}
\\
\multicolumn{4}{|c|}{FRRN~\cite{chen2018deeplab}}
&\multicolumn{4}{|c|}{71.8}&\multicolumn{2}{|c|}{98.2}
&\multicolumn{2}{|c|}{83.3}&\multicolumn{2}{|c|}{91.6}
&\multicolumn{2}{|c|}{45.8}&\multicolumn{2}{|c|}{51.1}
&\multicolumn{2}{|c|}{62.2}&\multicolumn{2}{|c|}{69.4}
&\multicolumn{2}{|c|}{72.4}&\multicolumn{2}{|c|}{92.6}
&\multicolumn{2}{|c|}{70.0}&\multicolumn{2}{|c|}{94.9}
&\multicolumn{2}{|c|}{81.6}&\multicolumn{2}{|c|}{62.7}
&\multicolumn{2}{|c|}{94.6}&\multicolumn{2}{|c|}{49.1}
&\multicolumn{2}{|c|}{67.1}&\multicolumn{2}{|c|}{55.3}
&\multicolumn{2}{|c|}{53.5}&\multicolumn{2}{|c|}{69.5}
\\
\multicolumn{4}{|c|}{RefineNet~\cite{lin2017refinenet}}
&\multicolumn{4}{|c|}{73.6}&\multicolumn{2}{|c|}{98.2}
&\multicolumn{2}{|c|}{83.3}&\multicolumn{2}{|c|}{91.3}
&\multicolumn{2}{|c|}{47.8}&\multicolumn{2}{|c|}{50.4}
&\multicolumn{2}{|c|}{56.1}&\multicolumn{2}{|c|}{66.9}
&\multicolumn{2}{|c|}{71.3}&\multicolumn{2}{|c|}{92.3}
&\multicolumn{2}{|c|}{70.3}&\multicolumn{2}{|c|}{94.8}
&\multicolumn{2}{|c|}{80.9}&\multicolumn{2}{|c|}{63.3}
&\multicolumn{2}{|c|}{94.5}&\multicolumn{2}{|c|}{64.6}
&\multicolumn{2}{|c|}{76.1}&\multicolumn{2}{|c|}{64.3}
&\multicolumn{2}{|c|}{62.2}&\multicolumn{2}{|c|}{70.0}
\\
\multicolumn{4}{|c|}{PEARL~\cite{jin2017video}}
&\multicolumn{4}{|c|}{75.4}&\multicolumn{2}{|c|}{98.4}
&\multicolumn{2}{|c|}{84.5}&\multicolumn{2}{|c|}{92.1}
&\multicolumn{2}{|c|}{54.1}&\multicolumn{2}{|c|}{56.6}
&\multicolumn{2}{|c|}{60.4}&\multicolumn{2}{|c|}{69.0}
&\multicolumn{2}{|c|}{74.1}&\multicolumn{2}{|c|}{92.9}
&\multicolumn{2}{|c|}{70.9}&\multicolumn{2}{|c|}{95.2}
&\multicolumn{2}{|c|}{83.5}&\multicolumn{2}{|c|}{65.7}
&\multicolumn{2}{|c|}{95.0}&\multicolumn{2}{|c|}{61.8}
&\multicolumn{2}{|c|}{72.2}&\multicolumn{2}{|c|}{69.6}
&\multicolumn{2}{|c|}{64.8}&\multicolumn{2}{|c|}{72.8}
\\
\multicolumn{4}{|c|}{DUC~\cite{wang2018understanding}}
&\multicolumn{4}{|c|}{77.6}&\multicolumn{2}{|c|}{98.5}
&\multicolumn{2}{|c|}{85.5}&\multicolumn{2}{|c|}{92.8}
&\multicolumn{2}{|c|}{58.6}&\multicolumn{2}{|c|}{55.5}
&\multicolumn{2}{|c|}{65.0}&\multicolumn{2}{|c|}{73.5}
&\multicolumn{2}{|c|}{77.9}&\multicolumn{2}{|c|}{93.3}
&\multicolumn{2}{|c|}{72.0}&\multicolumn{2}{|c|}{95.2}
&\multicolumn{2}{|c|}{84.8}&\multicolumn{2}{|c|}{68.5}
&\multicolumn{2}{|c|}{95.4}&\multicolumn{2}{|c|}{70.9}
&\multicolumn{2}{|c|}{78.8}&\multicolumn{2}{|c|}{68.7}
&\multicolumn{2}{|c|}{65.9}&\multicolumn{2}{|c|}{73.8}
\\
\multicolumn{4}{|c|}{PSPNet~\cite{zhao2017pyramid}}
&\multicolumn{4}{|c|}{78.4}&\multicolumn{2}{|c|}{98.6}
&\multicolumn{2}{|c|}{86.2}&\multicolumn{2}{|c|}{92.9}
&\multicolumn{2}{|c|}{50.8}&\multicolumn{2}{|c|}{58.8}
&\multicolumn{2}{|c|}{64.0}&\multicolumn{2}{|c|}{75.6}
&\multicolumn{2}{|c|}{79.0}&\multicolumn{2}{|c|}{93.4}
&\multicolumn{2}{|c|}{72.3}&\multicolumn{2}{|c|}{95.4}
&\multicolumn{2}{|c|}{86.5}&\multicolumn{2}{|c|}{71.3}
&\multicolumn{2}{|c|}{95.9}&\multicolumn{2}{|c|}{68.2}
&\multicolumn{2}{|c|}{79.5}&\multicolumn{2}{|c|}{73.8}
&\multicolumn{2}{|c|}{69.5}&\multicolumn{2}{|c|}{77.2}
\\
\multicolumn{4}{|c|}{PSPNet$^\ddagger$~\cite{zhao2017pyramid}}
&\multicolumn{4}{|c|}{80.2}&\multicolumn{2}{|c|}{98.6}
&\multicolumn{2}{|c|}{86.6}&\multicolumn{2}{|c|}{93.2}
&\multicolumn{2}{|c|}{58.1}&\multicolumn{2}{|c|}{63.0}
&\multicolumn{2}{|c|}{64.5}&\multicolumn{2}{|c|}{75.2}
&\multicolumn{2}{|c|}{79.2}&\multicolumn{2}{|c|}{93.4}
&\multicolumn{2}{|c|}{72.1}&\multicolumn{2}{|c|}{95.1}
&\multicolumn{2}{|c|}{86.3}&\multicolumn{2}{|c|}{71.4}
&\multicolumn{2}{|c|}{96.0}&\multicolumn{2}{|c|}{73.5}
&\multicolumn{2}{|c|}{90.4}&\multicolumn{2}{|c|}{80.3}
&\multicolumn{2}{|c|}{69.9}&\multicolumn{2}{|c|}{76.9}
\\
\multicolumn{4}{|c|}{WResNet~\cite{wu2016wider}}
&\multicolumn{4}{|c|}{78.4}&\multicolumn{2}{|c|}{98.5}
&\multicolumn{2}{|c|}{85.7}&\multicolumn{2}{|c|}{93.1}
&\multicolumn{2}{|c|}{55.5}&\multicolumn{2}{|c|}{59.1}
&\multicolumn{2}{|c|}{67.1}&\multicolumn{2}{|c|}{74.8}
&\multicolumn{2}{|c|}{78.7}&\multicolumn{2}{|c|}{93.7}
&\multicolumn{2}{|c|}{72.6}&\multicolumn{2}{|c|}{95.5}
&\multicolumn{2}{|c|}{86.6}&\multicolumn{2}{|c|}{69.2}
&\multicolumn{2}{|c|}{95.7}&\multicolumn{2}{|c|}{64.5}
&\multicolumn{2}{|c|}{78.8}&\multicolumn{2}{|c|}{74.1}
&\multicolumn{2}{|c|}{69.0}&\multicolumn{2}{|c|}{76.7}
\\
\multicolumn{4}{|c|}{DenseASPP$\ddagger$~\cite{yang2018denseaspp}}
&\multicolumn{4}{|c|}{80.6}&\multicolumn{2}{|c|}{98.7}
&\multicolumn{2}{|c|}{87.1}&\multicolumn{2}{|c|}{93.4}
&\multicolumn{2}{|c|}{60.7}&\multicolumn{2}{|c|}{62.7}
&\multicolumn{2}{|c|}{65.6}&\multicolumn{2}{|c|}{74.6}
&\multicolumn{2}{|c|}{78.5}&\multicolumn{2}{|c|}{93.6}
&\multicolumn{2}{|c|}{72.5}&\multicolumn{2}{|c|}{95.4}
&\multicolumn{2}{|c|}{86.2}&\multicolumn{2}{|c|}{71.9}
&\multicolumn{2}{|c|}{96.0}&\multicolumn{2}{|c|}{78.0}
&\multicolumn{2}{|c|}{90.3}&\multicolumn{2}{|c|}{80.7}
&\multicolumn{2}{|c|}{69.7}&\multicolumn{2}{|c|}{76.8}
\\
\multicolumn{4}{|c|}{GANet$\ddagger$~\cite{zhang2019deep}}
&\multicolumn{4}{|c|}{82.8}&\multicolumn{2}{|c|}{99.4}
&\multicolumn{2}{|c|}{88.1}&\multicolumn{2}{|c|}{94.5}
&\multicolumn{2}{|c|}{62.4}&\multicolumn{2}{|c|}{65.3}
&\multicolumn{2}{|c|}{69.5}&\multicolumn{2}{|c|}{77.3}
&\multicolumn{2}{|c|}{81.3}&\multicolumn{2}{|c|}{95.6}
&\multicolumn{2}{|c|}{75.4}&\multicolumn{2}{|c|}{97.6}
&\multicolumn{2}{|c|}{88.6}&\multicolumn{2}{|c|}{74.9}
&\multicolumn{2}{|c|}{97.3}&\multicolumn{2}{|c|}{79.2}
&\multicolumn{2}{|c|}{92.5}&\multicolumn{2}{|c|}{82.3}
&\multicolumn{2}{|c|}{72.5}&\multicolumn{2}{|c|}{78.6}
\\
\multicolumn{4}{|c|}{DMLS$\ddagger$ (Ours)}
&\multicolumn{4}{|c|}{\textbf{83.7}}&\multicolumn{2}{|c|}{\textbf{99.6}}
&\multicolumn{2}{|c|}{\textbf{88.9}}&\multicolumn{2}{|c|}{\textbf{95.2}}
&\multicolumn{2}{|c|}{\textbf{63.7}}&\multicolumn{2}{|c|}{\textbf{67.9}}
&\multicolumn{2}{|c|}{\textbf{70.8}}&\multicolumn{2}{|c|}{\textbf{78.4}}
&\multicolumn{2}{|c|}{\textbf{82.6}}&\multicolumn{2}{|c|}{\textbf{96.3}}
&\multicolumn{2}{|c|}{\textbf{76.8}}&\multicolumn{2}{|c|}{\textbf{98.2}}
&\multicolumn{2}{|c|}{\textbf{89.1}}&\multicolumn{2}{|c|}{\textbf{75.3}}
&\multicolumn{2}{|c|}{\textbf{98.1}}&\multicolumn{2}{|c|}{\textbf{80.5}}
&\multicolumn{2}{|c|}{\textbf{92.7}}&\multicolumn{2}{|c|}{\textbf{83.1}}
&\multicolumn{2}{|c|}{\textbf{73.4}}&\multicolumn{2}{|c|}{\textbf{79.8}}
\\
\hline
\end{tabular}
}
\vspace{-4mm}
\label{table:cityscapes_test_cls}
\end{center}
\end{table*}
\begin{figure*}
\centering
\begin{tabular}{@{}c@{}c@{}c@{}c@{}c@{}c@{}c@{}c@{}c@{}c@{}c@{}c@{}c@{}c@{}c@{}c@{}c@{}c@{}c}
(a)\hspace{3cm}(b)\hspace{3cm}(c)\hspace{3cm}(d)\hspace{3cm}(e)\\
\includegraphics[width=\linewidth,height=6cm]{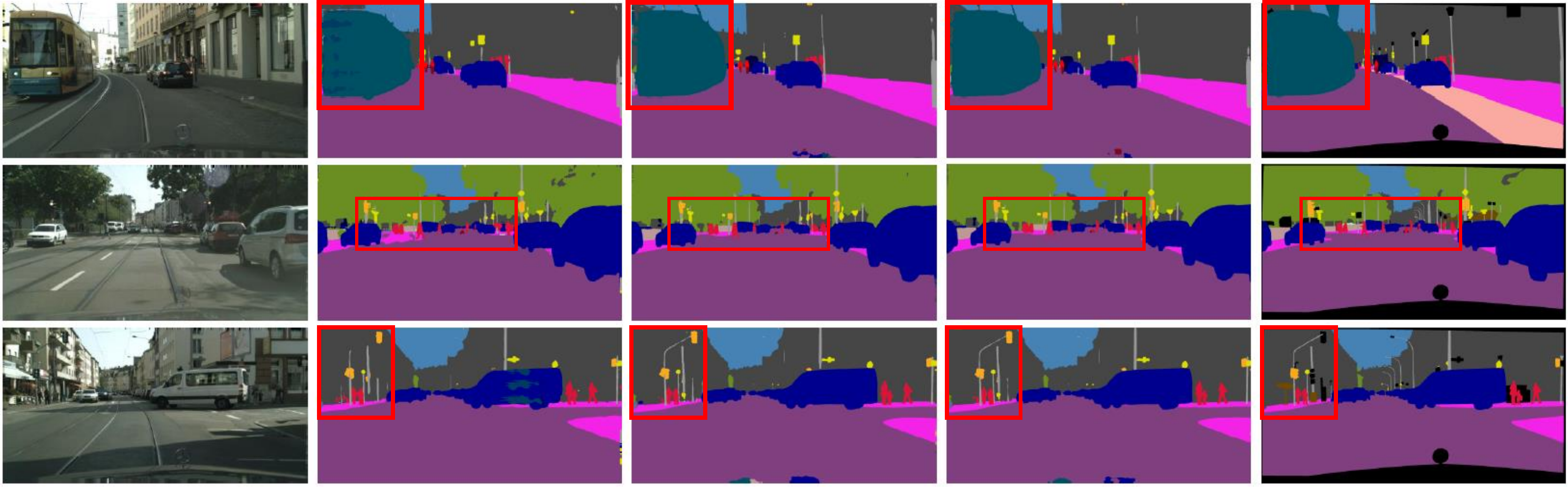} \\
\end{tabular}
\caption{Typical parsing results on the Cityscapes dataset. From left to right: (a) Input Images, (b) PSPNet$^{\ddagger}$(ResNet-101)~\cite{zhao2017pyramid}, (c) DeeplabV3+$^{\ddagger}$ (ResNet-101)~\cite{chen2018encoder}, (d) Ours, (e) Ground Truth (ResNet-101).
Our results contain more detailed boundary structures than other state-of-the-art baselines.}
\vspace{-4mm}
\label{fig:cityscapes_comparison}
\end{figure*}
\subsection{Evaluation on Cityscapes Dataset}
The Cityscapes dataset~\cite{cordts2016cityscapes} is a common reference for street scene parsing due to its highly varied scenarios and challenging labeled classes.
It contains 5,000 images with fine annotations, divided to 2,975 images for training, 500 images for validation and 1,525 images for testing.
Labels of the test set are not available, but it is possible to evaluate them on the online test server.
There are also available 20K coarsely annotated images, which can be used to pre-train the models.

To train our model, we merge the training and validation sets with fine annotations.
For fair comparison, we follow most existing works, and adopt the widely-used 19 categories.
We also report the parsing results with different FCN backbones.
Tab.~\ref{table:cityscapes_test} shows the quantitative results on the test set.
It can be observed that our methods perform better than other methods with notable advantages.
For example, with the same VGG-16 backbone~\cite{simonyan2014very}, our approach achieves better results than other competitors, even trained with the additional coarse data, such as LRR~\cite{ghiasi2016laplacian} and CNN+CRF~\cite{lin2016efficient}.
With the ResNet-50 backbone~\cite{he2016deep}, our approach performs better than most of the compared methods, especially the PSPNet~\cite{zhao2017pyramid} which utilizes a much deeper ResNet (101-layers).
With the more powerful ResNet-101 backbone, our method consistently improves the parsing performance.
Compared with the very recent Deeplab v3+~\cite{chen2018encoder} and GANet~\cite{zhang2019deep}, our approach universally boosts the performance about 1\% in terms of near all metrics.
As shown in Tab.~\ref{table:cityscapes_test}, using both fine and coarse data for training brings benefit to all models.
With fine and coarse data, our ResNet-101-based model yields expressive performances, \emph{i.e.}, 83.7 class IoU, 66.1 category IoU, 93.0 instance-level class IoU, and 83.3 instance-level category IoU.
Tab.~\ref{table:cityscapes_test_cls} show the category-wise quantitative results of representative methods.

Several typical examples are shown in Fig.~\ref{fig:cityscapes_comparison}.
It can be observed that, our results constantly contain more accurate and detailed structures compared to the baselines.
The main reason is the introduction of MLS for boundary refinements.
%
%
\subsection{Evaluation on Mapillary Vistas Dataset}
\begin{table}
\begin{center}
\caption{Quantitative results on the Mapillary Vistas validation dataset. The best two results are in bold and underline.}
\resizebox{0.4\textwidth}{!}
{
\begin{tabular}{c|c|ccccc}
\hline
Methods &FCN backbone&Mean IoU\\
\hline
Baseline FCN8s~\cite{long2015fully} &ResNet-101&38.48\\
WResNet~\cite{wu2016wider}& ResNet-38 &41.12        \\
Loss max-pooling~\cite{rota2017loss}& ResNet-38  &43.78 \\
PSPNet~\cite{zhao2017pyramid}&ResNet-101&49.76\\
PSPNet (4 ensemble)&ResNet-101&53.85\\
Inplace-ABN~\cite{rota2018place}&ResNet-38&53.12\\
IIAU-Adelaide~\cite{zhang2017amulet}&VGG-19&42.25\\
DSSPN~\cite{liang2018dynamic}&ResNet-101&45.01\\
GANet~\cite{zhang2019deep} &VGG-16&41.47\\
GANet~\cite{zhang2019deep} &ResNet-50&53.04\\
GANet~\cite{zhang2019deep} &ResNet-101&\underline{54.21}\\
\hline
DMLS (Ours)&VGG-16 & 41.71 \\
DMLS (Ours)&ResNet-50&54.42\\
DMLS (Ours)&ResNet-101&\textbf{54.83}\\
\hline
\end{tabular}
}
\vspace{-4mm}
\label{table:mvd_cls}
\end{center}
\end{table}
The Mapillary Vistas dataset~\cite{neuhold2017mapillary} has been collected to cover the diversity in street appearances.
It comprises 25k densely annotated images, which are split into 18k/2k/5k images for training, validation and testing respectively.
The number of semantic categories increases to 66, which makes the parsing task more challenging.
Thus, there are only few works performed experiments on this dataset.
Besides, the testing server only opens at specific time.
Based on this fact, we follow previous works~\cite{zhao2017pyramid,rota2018place,liang2018dynamic}, and mainly report the parsing performance on the validation dataset.

Tab.~\ref{table:mvd_cls} presents quantitative results with other state-of-the-art methods.
For fair comparison, we also re-implement the FCN8s baseline~\cite{long2015fully} with the ResNet-101 backbone.
Among the existing methods, the recent GANet~\cite{zhang2019deep} achieves best results, even better than the PSPNet with the multi-model ensemble.
While our method only with the ResNet-50 backbone achieves a competing result to the GANet.
With the ResNet-101 backbone, our method achieves best results on this dataset.
We also observe that our model with the VGG-16 backbone performs better than the ResNet-38 based WResNet~\cite{wu2016wider} and the ResNet-101 based FCN8s.
They adopt a deeper network for feature extractions.
This further verifies the effectiveness of our methods, even simple backbone networks are used.
\subsection{Evaluation on ADE20K Dataset}
\begin{figure*}
\centering
\resizebox{1\textwidth}{!}
{
\begin{tabular}{@{}c@{}c@{}c@{}c@{}c@{}c@{}c@{}c@{}c@{}c@{}c@{}c@{}c@{}c@{}c@{}c@{}c@{}c@{}c}
{\small(a)} & {\small(b)} & {\small(c)} & {\small(d)} & {\small(e)} & {\small(f)}\\
\vspace{-0.5mm}
\includegraphics[width=0.15\linewidth,height=1.45cm]{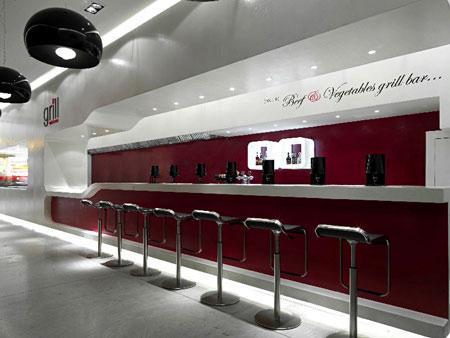}&
\includegraphics[width=0.15\linewidth,height=1.45cm]{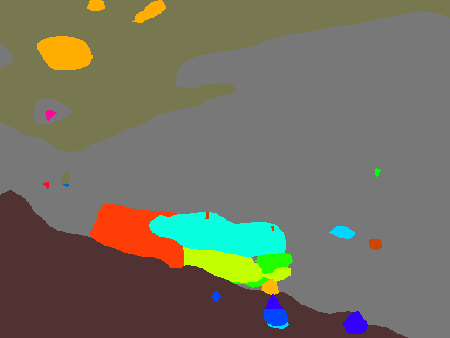} &
\includegraphics[width=0.15\linewidth,height=1.45cm]{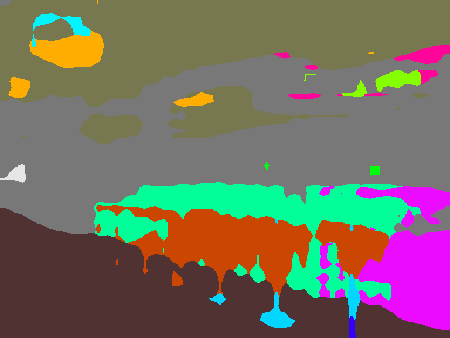} &
\includegraphics[width=0.15\linewidth,height=1.45cm]{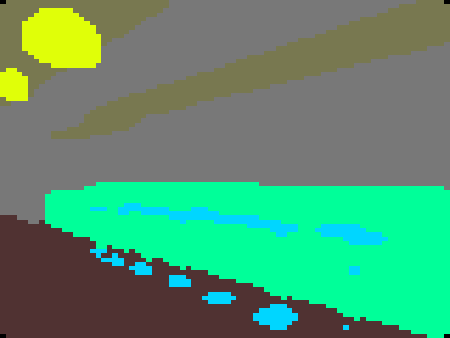} &
\includegraphics[width=0.15\linewidth,height=1.45cm]{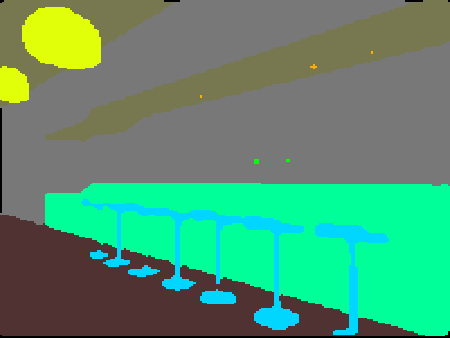} &
\includegraphics[width=0.15\linewidth,height=1.45cm]{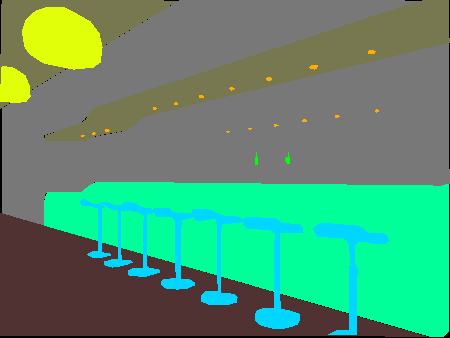} \\
\vspace{-0.5mm}
\includegraphics[width=0.15\linewidth,height=1.45cm]{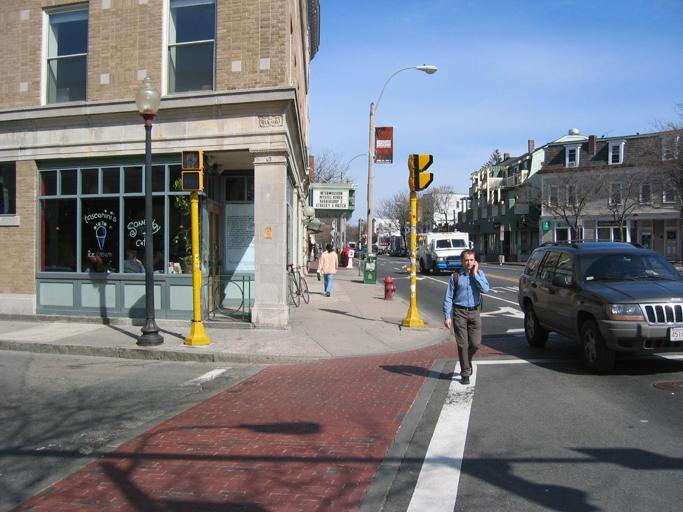}&
\includegraphics[width=0.15\linewidth,height=1.45cm]{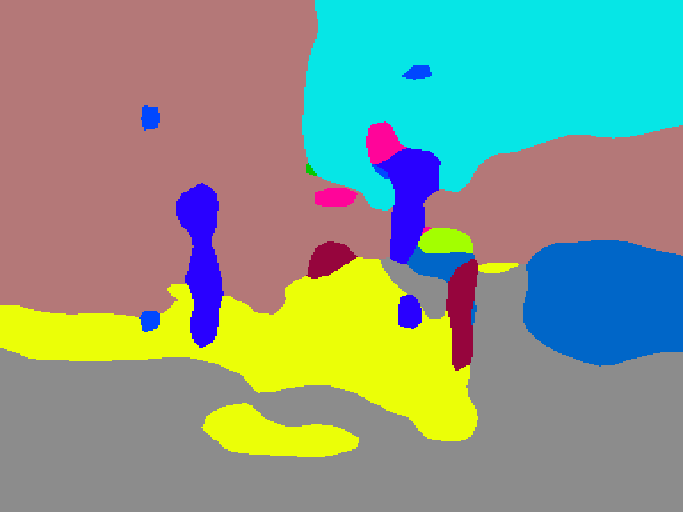} &
\includegraphics[width=0.15\linewidth,height=1.45cm]{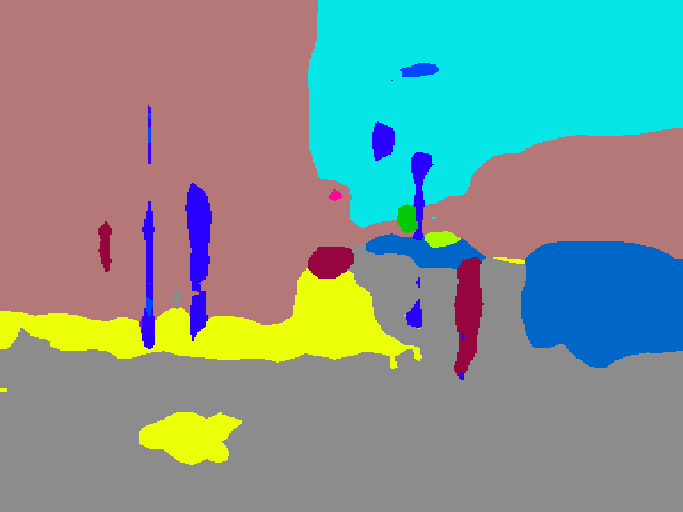} &
\includegraphics[width=0.15\linewidth,height=1.45cm]{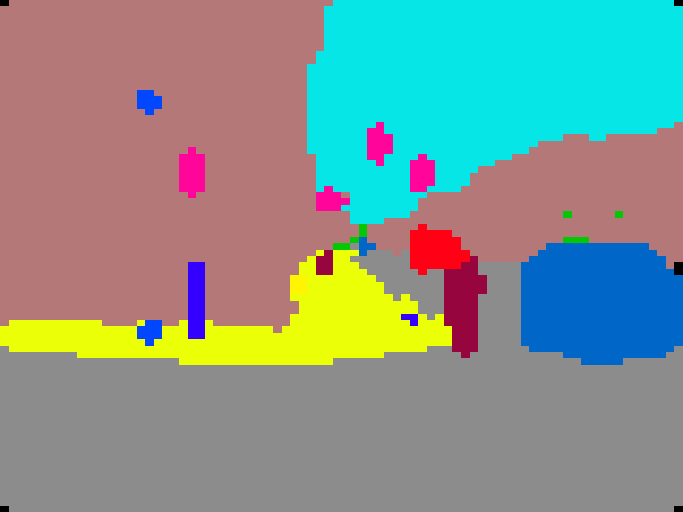} &
\includegraphics[width=0.15\linewidth,height=1.45cm]{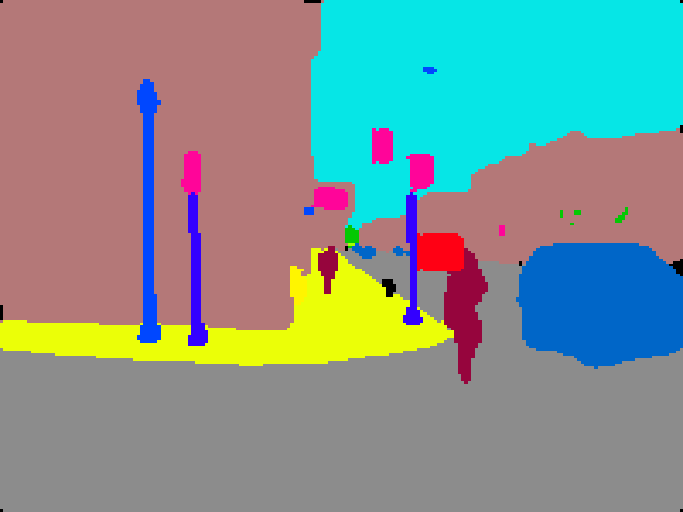} &
\includegraphics[width=0.15\linewidth,height=1.45cm]{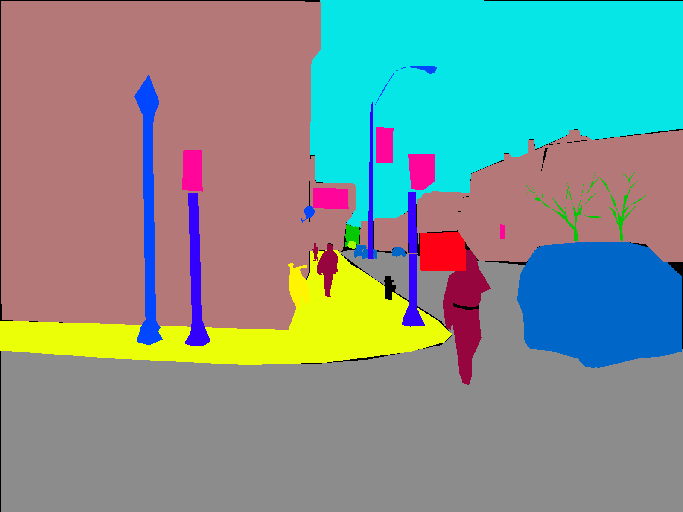} \\
\vspace{-0.5mm}
\includegraphics[width=0.15\linewidth,height=1.45cm]{imgs_ADE_val_00000203.jpg}&
\includegraphics[width=0.15\linewidth,height=1.45cm]{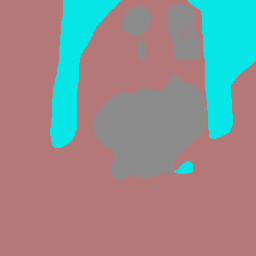} &
\includegraphics[width=0.15\linewidth,height=1.45cm]{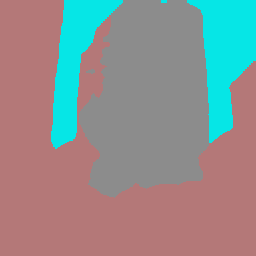} &
\includegraphics[width=0.15\linewidth,height=1.45cm]{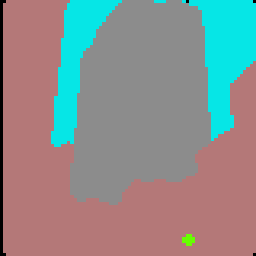} &
\includegraphics[width=0.15\linewidth,height=1.45cm]{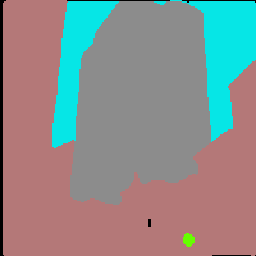} &
\includegraphics[width=0.15\linewidth,height=1.45cm]{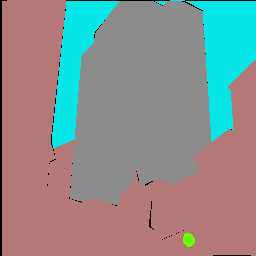} \\
\vspace{-0.5mm}
\includegraphics[width=0.15\linewidth,height=1.45cm]{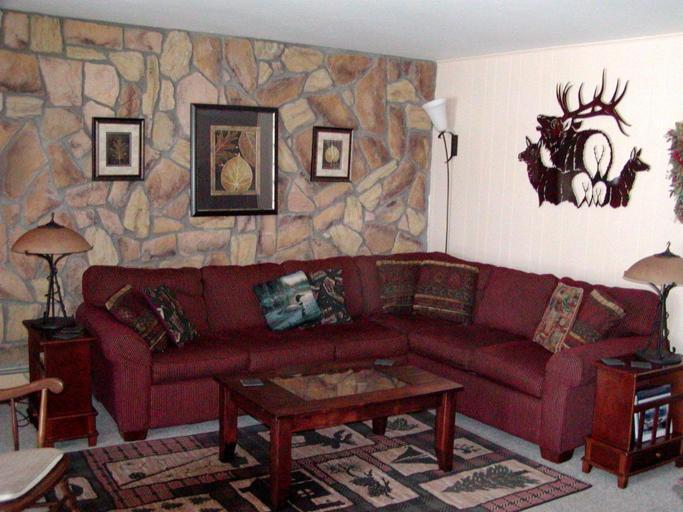}&
\includegraphics[width=0.15\linewidth,height=1.45cm]{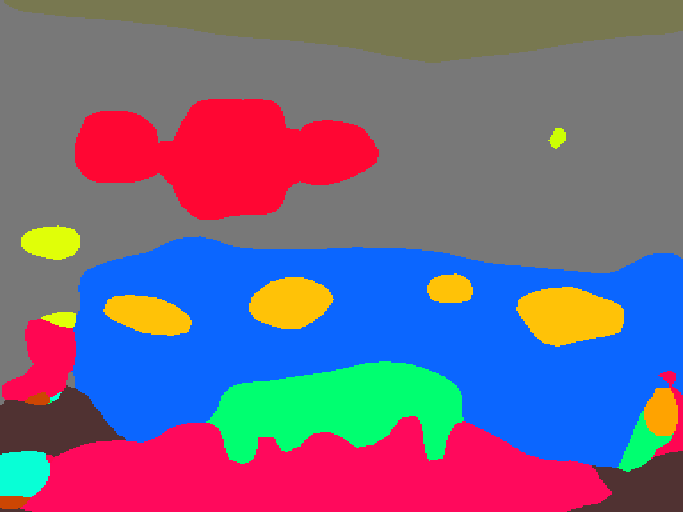} &
\includegraphics[width=0.15\linewidth,height=1.45cm]{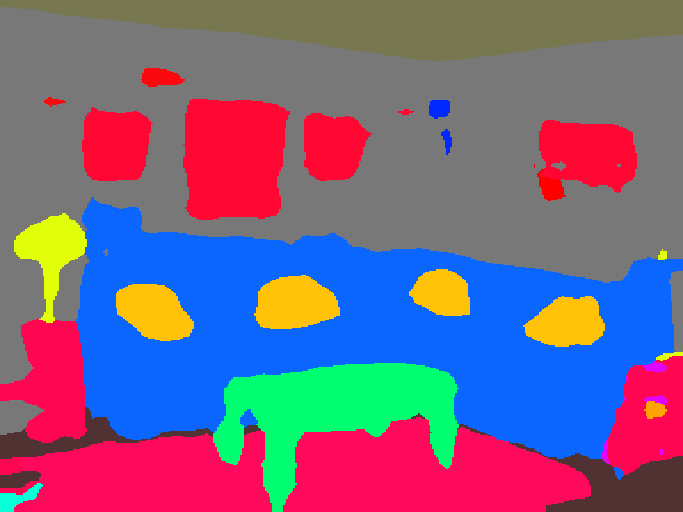} &
\includegraphics[width=0.15\linewidth,height=1.45cm]{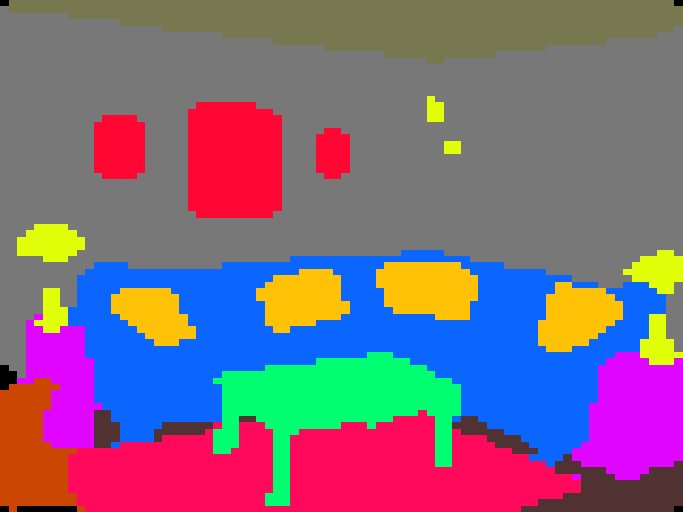} &
\includegraphics[width=0.15\linewidth,height=1.45cm]{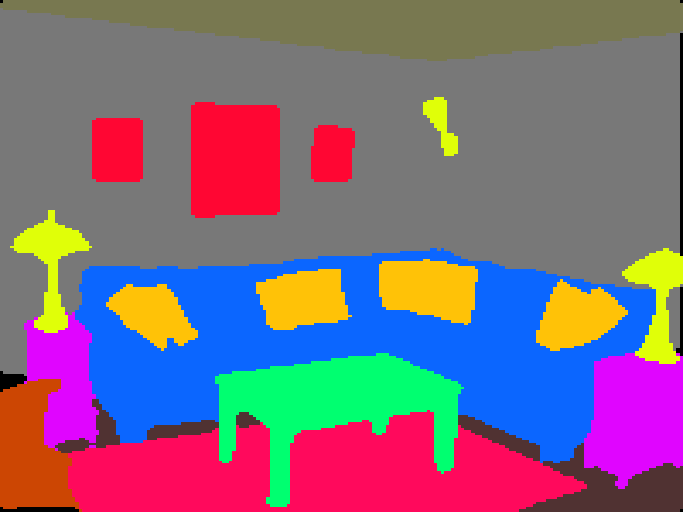} &
\includegraphics[width=0.15\linewidth,height=1.45cm]{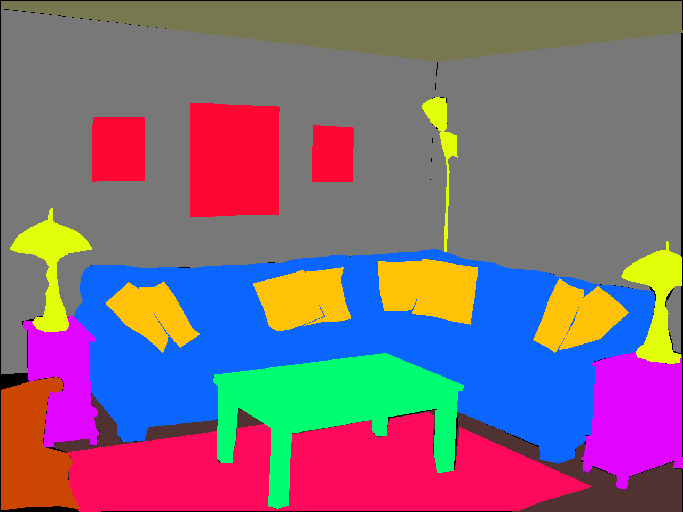} \\
\vspace{-0.5mm}
\includegraphics[width=0.15\linewidth,height=1.45cm]{imgs_ADE_val_00000832.jpg}&
\includegraphics[width=0.15\linewidth,height=1.45cm]{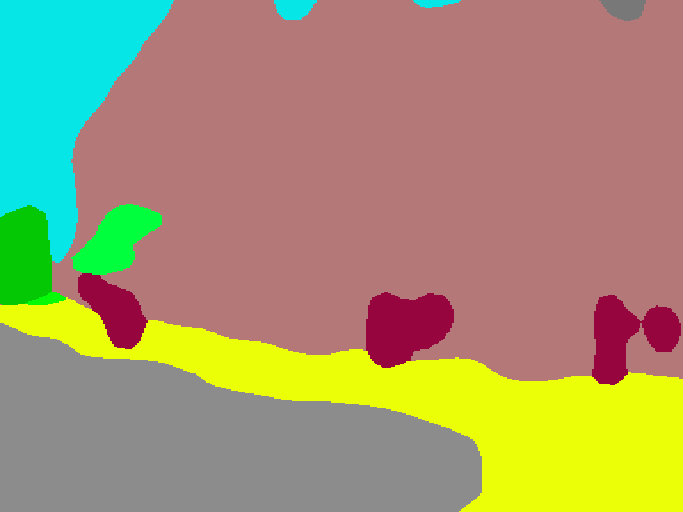} &
\includegraphics[width=0.15\linewidth,height=1.45cm]{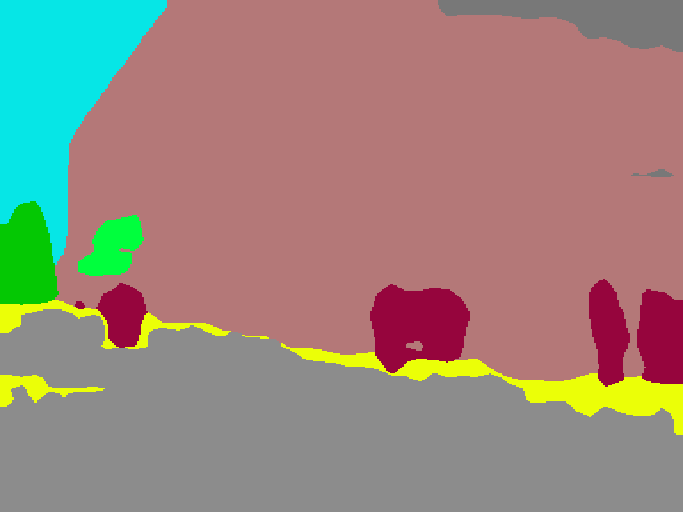} &
\includegraphics[width=0.15\linewidth,height=1.45cm]{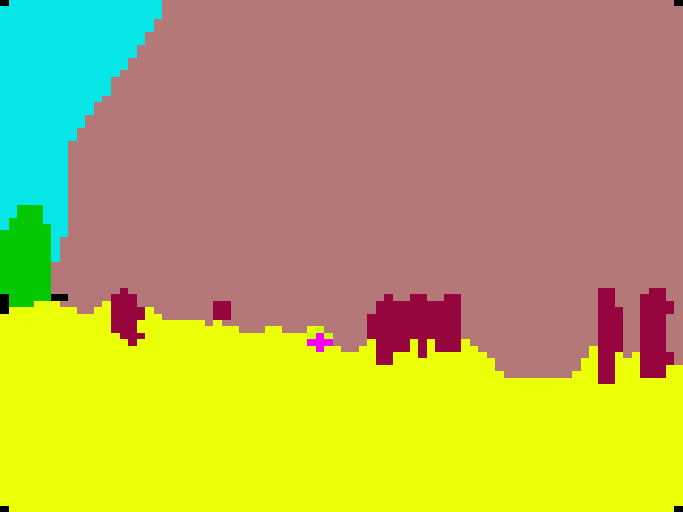} &
\includegraphics[width=0.15\linewidth,height=1.45cm]{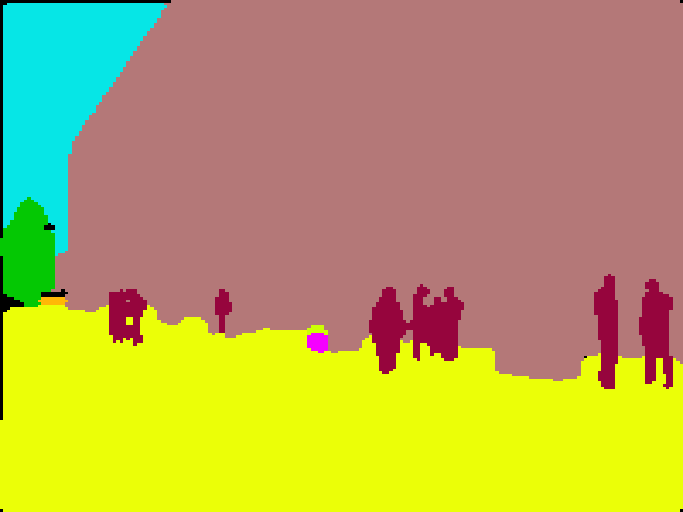} &
\includegraphics[width=0.15\linewidth,height=1.45cm]{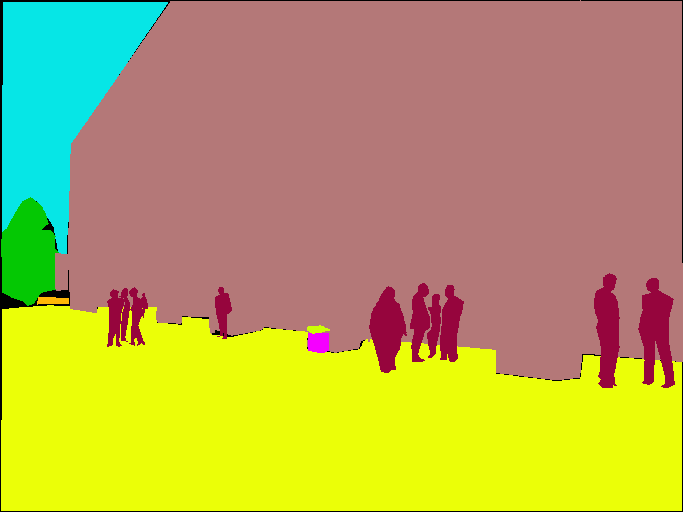} \\
\end{tabular}
}
\caption{Typical parsing results on the ADE20K dataset. From left to right: (a) Input Images, (b) FCN8s (ResNet-101)~\cite{long2015fully}, (c) DilatedNet~\cite{yu2016multi}, (d) PSPNet~\cite{zhao2017pyramid}, (e) Ours (ResNet-101); (f) Ground Truth.}
\label{fig:ade20k_comparison}
\end{figure*}
\begin{table}
\begin{center}
\doublerulesep=0.5pt
\caption{Quantitative results on the ADE20K validation set. The best two results are in bold and underline.}
\resizebox{0.45\textwidth}{!}
{
\begin{tabular}{|c|c|c|c|c|c|c|c|c|c|c|c|c|c|c|c|c|c|c|c|c|c|c|c|c|c|c|c|c|||c|c|c|c|c|c|c|c|}
\hline
\multicolumn{4}{c|}{Methods} &\multicolumn{4}{|c|}{FCN backbone} &\multicolumn{2}{|c|}{Pixel Acc.\%}&\multicolumn{2}{|c}{Mean IoU\%}
\\
\hline
\multicolumn{4}{c|}{FCN8s~\cite{long2015fully}}&\multicolumn{4}{|c|}{VGG-16}
&\multicolumn{2}{|c|}{71.32}&\multicolumn{2}{|c}{29.39}
\\
\multicolumn{4}{c|}{FCN8s~\cite{long2015fully}}&\multicolumn{4}{|c|}{ResNet-50}
&\multicolumn{2}{|c|}{74.57}&\multicolumn{2}{|c}{34.38}
\\
\multicolumn{4}{c|}{FCN8s~\cite{long2015fully}}&\multicolumn{4}{|c|}{ResNet-101}
&\multicolumn{2}{|c|}{74.83}&\multicolumn{2}{|c}{34.62}
\\
\multicolumn{4}{c|}{DilatedNet~\cite{yu2016multi}}&\multicolumn{4}{|c|}{VGG-16}
&\multicolumn{2}{|c|}{73.55}&\multicolumn{2}{|c}{32.31}
\\
\multicolumn{4}{c|}{SegNet~\cite{badrinarayanan2017segnet}}&\multicolumn{4}{|c|}{VGG-16}
&\multicolumn{2}{|c|}{71.00}&\multicolumn{2}{|c}{21.64}
\\
\multicolumn{4}{c|}{CascadeNet~\cite{zhou2017scene}}&\multicolumn{4}{|c|}{VGG-16}
&\multicolumn{2}{|c|}{74.52}&\multicolumn{2}{|c}{34.90}
\\
\multicolumn{4}{c|}{RefineNet~\cite{lin2017refinenet}}&\multicolumn{4}{|c|}{ResNet-101}
&\multicolumn{2}{|c|}{--}&\multicolumn{2}{|c}{40.20}
\\
\multicolumn{4}{c|}{RefineNet~\cite{lin2017refinenet}}&\multicolumn{4}{|c|}{ResNet-152}
&\multicolumn{2}{|c|}{--}&\multicolumn{2}{|c}{40.71}
\\
\multicolumn{4}{c|}{PSPNet~\cite{zhao2017pyramid}}&\multicolumn{4}{|c|}{ResNet-50}
&\multicolumn{2}{|c|}{80.04}&\multicolumn{2}{|c}{41.68}
\\
\multicolumn{4}{c|}{PSPNet~\cite{zhao2017pyramid}}&\multicolumn{4}{|c|}{ResNet-101}
&\multicolumn{2}{|c|}{81.39}&\multicolumn{2}{|c}{43.29}
\\
\multicolumn{4}{c|}{PSPNet~\cite{zhao2017pyramid}}&\multicolumn{4}{|c|}{ResNet-269}
&\multicolumn{2}{|c|}{81.69}&\multicolumn{2}{|c}{44.94}
\\
\multicolumn{4}{c|}{SAC~\cite{zhang2017scale}}&\multicolumn{4}{|c|}{ResNet-101}
&\multicolumn{2}{|c|}{81.86}&\multicolumn{2}{|c}{44.30}
\\
\multicolumn{4}{c|}{DeepLabv2~\cite{chen2018deeplab}}&\multicolumn{4}{|c|}{VGG-16}
&\multicolumn{2}{|c|}{74.88}&\multicolumn{2}{|c}{33.03}
\\
\multicolumn{4}{c|}{DeepLabv2~\cite{chen2018deeplab}}&\multicolumn{4}{|c|}{ResNet-101}
&\multicolumn{2}{|c|}{77.31}&\multicolumn{2}{|c}{36.75}
\\
\multicolumn{4}{c|}{MoE-SPNet~\cite{fu2018moe}}&\multicolumn{4}{|c|}{VGG-16}
&\multicolumn{2}{|c|}{75.50}&\multicolumn{2}{|c}{34.35}
\\
\multicolumn{4}{c|}{MoE-SPNet~\cite{fu2018moe}}&\multicolumn{4}{|c|}{ResNet-101}
&\multicolumn{2}{|c|}{78.02}&\multicolumn{2}{|c}{37.89}
\\
\multicolumn{4}{c|}{EncNet~\cite{zhang2018context}}&\multicolumn{4}{|c|}{ResNet-50}
&\multicolumn{2}{|c|}{79.73}&\multicolumn{2}{|c}{41.11}
\\
\multicolumn{4}{c|}{EncNet~\cite{zhang2018context}}&\multicolumn{4}{|c|}{ResNet-101}
&\multicolumn{2}{|c|}{81.69}&\multicolumn{2}{|c}{44.65}
\\
\multicolumn{4}{c|}{DSSPN~\cite{liang2018dynamic}}&\multicolumn{4}{|c|}{VGG-16}
&\multicolumn{2}{|c|}{76.04}&\multicolumn{2}{|c}{34.56}
\\
\multicolumn{4}{c|}{DSSPN~\cite{liang2018dynamic}}&\multicolumn{4}{|c|}{ResNet-101}
&\multicolumn{2}{|c|}{81.13}&\multicolumn{2}{|c}{43.68}
\\
\multicolumn{4}{c|}{PSANet~\cite{zhao2018psanet}}&\multicolumn{4}{|c|}{ResNet-50}
&\multicolumn{2}{|c|}{80.92}&\multicolumn{2}{|c}{42.97}
\\
\multicolumn{4}{c|}{PSANet~\cite{zhao2018psanet}}&\multicolumn{4}{|c|}{ResNet-101}
&\multicolumn{2}{|c|}{81.51}&\multicolumn{2}{|c}{43.77}
\\
\multicolumn{4}{c|}{UperNet~\cite{xiao2018unified}}&\multicolumn{4}{|c|}{ResNet-50}
&\multicolumn{2}{|c|}{79.98}&\multicolumn{2}{|c}{41.22}
\\
\multicolumn{4}{c|}{GANet~\cite{zhang2019deep}}&\multicolumn{4}{|c|}{VGG-16}
&\multicolumn{2}{|c|}{78.64}&\multicolumn{2}{|c}{41.52}
\\
\multicolumn{4}{c|}{GANet~\cite{zhang2019deep}}&\multicolumn{4}{|c|}{ResNet-50}
&\multicolumn{2}{|c|}{81.80}&\multicolumn{2}{|c}{44.71}
\\
\multicolumn{4}{c|}{GANet~\cite{zhang2019deep}}&\multicolumn{4}{|c|}{ResNet-101}
&\multicolumn{2}{|c|}{82.14}&\multicolumn{2}{|c}{45.36}
\\
\multicolumn{4}{c|}{WResNet~\cite{wu2016wider}}&\multicolumn{4}{|c|}{ResNet-101}
&\multicolumn{2}{|c|}{81.17}&\multicolumn{2}{|c}{43.73}
\\
\hline
\multicolumn{4}{c|}{OCNet~\cite{yuan2018ocnet}}&\multicolumn{4}{|c|}{ResNet-50}
&\multicolumn{2}{|c|}{80.44}&\multicolumn{2}{|c}{42.63}
\\
\multicolumn{4}{c|}{OCNet~\cite{yuan2018ocnet}}&\multicolumn{4}{|c|}{ResNet-101}
&\multicolumn{2}{|c|}{81.86}&\multicolumn{2}{|c}{45.08}
\\
\multicolumn{4}{c|}{Fast-OCNet~\cite{yuan2018ocnet}}&\multicolumn{4}{|c|}{ResNet-101}
&\multicolumn{2}{|c|}{82.03}&\multicolumn{2}{|c}{45.45}
\\
\hline
\multicolumn{4}{c|}{DMLS (Ours)}&\multicolumn{4}{|c|}{VGG-16}
&\multicolumn{2}{|c|}{81.03}&\multicolumn{2}{|c}{42.85}
\\
\multicolumn{4}{c|}{DMLS (Ours)}&\multicolumn{4}{|c|}{ResNet-50}
&\multicolumn{2}{|c|}{\underline{82.76}}&\multicolumn{2}{|c}{\underline{45.52}}
\\
\multicolumn{4}{c|}{DMLS (Ours)}&\multicolumn{4}{|c|}{ResNet-101}
&\multicolumn{2}{|c|}{\textbf{83.20}}&\multicolumn{2}{|c}{\textbf{46.16}}
\\
\hline
\end{tabular}
}
\vspace{-4mm}
\label{table:ade20k_val}
\end{center}
\end{table}
%
\begin{table}
\begin{center}
\doublerulesep=0.5pt
\caption{Quantitative results on the ADE20K test set. The best two results are in bold and underline.}
\resizebox{0.45\textwidth}{!}
{
\begin{tabular}{|c|c|c|c|c|c|c|c|c|c|c|c|c|c|c|c|c|c|c|c|c|c|c|c|c|c|c|c|c|||c|c|c|c|c|c|c|c|}
\hline
\multicolumn{4}{c|}{Methods}&\multicolumn{4}{|c|}{FCN backbone} &\multicolumn{2}{|c|}{Pixel Acc.\%}&\multicolumn{2}{|c}{Mean IoU\%}
\\
\hline
\multicolumn{4}{c|}{FCN8s~\cite{long2015fully}}&\multicolumn{4}{|c|}{VGG-16}
&\multicolumn{2}{|c|}{64.77}&\multicolumn{2}{|c}{24.83}
\\
\multicolumn{4}{c|}{DilatedNet~\cite{yu2016multi}}&\multicolumn{4}{|c|}{VGG-16}
&\multicolumn{2}{|c|}{65.41}&\multicolumn{2}{|c}{25.92}
\\
\multicolumn{4}{c|}{SegNet~\cite{badrinarayanan2017segnet}}&\multicolumn{4}{|c|}{VGG-16}
&\multicolumn{2}{|c|}{64.03}&\multicolumn{2}{|c}{17.54}
\\
\multicolumn{4}{c|}{PSPNet~\cite{zhao2017pyramid}}&\multicolumn{4}{|c|}{ResNet-269}
&\multicolumn{2}{|c|}{74.78}&\multicolumn{2}{|c}{39.73}
\\
\multicolumn{4}{c|}{DRN~\cite{zhuang2018dense}}&\multicolumn{4}{|c|}{ResNet-101}
&\multicolumn{2}{|c|}{73.94}&\multicolumn{2}{|c}{38.77}
\\
\multicolumn{4}{c|}{EncNet~\cite{zhang2018context}}&\multicolumn{4}{|c|}{ResNet-101}
&\multicolumn{2}{|c|}{73.74}&\multicolumn{2}{|c}{38.17}
\\
\multicolumn{4}{c|}{GANet~\cite{zhang2019deep}}&\multicolumn{4}{|c|}{VGG-16}
&\multicolumn{2}{|c|}{67.38}&\multicolumn{2}{|c}{27.43}
\\
\multicolumn{4}{c|}{GANet~\cite{zhang2019deep}}&\multicolumn{4}{|c|}{ResNet-50}
&\multicolumn{2}{|c|}{74.54}&\multicolumn{2}{|c}{39.03}
\\
\multicolumn{4}{c|}{GANet~\cite{zhang2019deep}}&\multicolumn{4}{|c|}{ResNet-101}
&\multicolumn{2}{|c|}{75.04}&\multicolumn{2}{|c}{40.11}
\\
\hline
\multicolumn{4}{c|}{eaglevison}&\multicolumn{4}{|c|}{Unknown}
&\multicolumn{2}{|c|}{69.73}&\multicolumn{2}{|c}{29.71}
\\
\multicolumn{4}{c|}{rainbowsecret}&\multicolumn{4}{|c|}{Unknown}
&\multicolumn{2}{|c|}{71.16}&\multicolumn{2}{|c}{33.95}
\\
\multicolumn{4}{c|}{vaq007}&\multicolumn{4}{|c|}{Unknown}
&\multicolumn{2}{|c|}{71.60}&\multicolumn{2}{|c}{33.61}
\\
\multicolumn{4}{c|}{kongyawen}&\multicolumn{4}{|c|}{Unknown}
&\multicolumn{2}{|c|}{72.22}&\multicolumn{2}{|c}{35.33}
\\
\multicolumn{4}{c|}{JunjunHe}&\multicolumn{4}{|c|}{Unknown}
&\multicolumn{2}{|c|}{72.89}&\multicolumn{2}{|c}{38.45}
\\
\multicolumn{4}{c|}{MSNonlocal}&\multicolumn{4}{|c|}{Unknown}
&\multicolumn{2}{|c|}{72.28}&\multicolumn{2}{|c}{35.57}
\\
\multicolumn{4}{c|}{fromandto}&\multicolumn{4}{|c|}{Unknown}
&\multicolumn{2}{|c|}{72.39}&\multicolumn{2}{|c}{35.79}
\\
\multicolumn{4}{c|}{101Net}&\multicolumn{4}{|c|}{ResNet-101}
&\multicolumn{2}{|c|}{72.89}&\multicolumn{2}{|c}{38.03}
\\
\multicolumn{4}{c|}{360+MCG-ICT-CAS-SP}&\multicolumn{4}{|c|}{Unknown}
&\multicolumn{2}{|c|}{73.73}&\multicolumn{2}{|c}{37.50}
\\
\multicolumn{4}{c|}{DSN}&\multicolumn{4}{|c|}{Unknown}
&\multicolumn{2}{|c|}{73.85}&\multicolumn{2}{|c}{38.47}
\\
\multicolumn{4}{c|}{APCNet-SingleModel}&\multicolumn{4}{|c|}{Unknown}
&\multicolumn{2}{|c|}{72.94}&\multicolumn{2}{|c}{38.39}
\\
\multicolumn{4}{c|}{SSN}&\multicolumn{4}{|c|}{Unknown}
&\multicolumn{2}{|c|}{72.08}&\multicolumn{2}{|c}{35.78}
\\
\multicolumn{4}{c|}{CDN}&\multicolumn{4}{|c|}{Unknown}
&\multicolumn{2}{|c|}{72.11}&\multicolumn{2}{|c}{35.97}
\\
\multicolumn{4}{c|}{deepnet}&\multicolumn{4}{|c|}{Unknown}
&\multicolumn{2}{|c|}{72.86}&\multicolumn{2}{|c}{36.57}
\\
\multicolumn{4}{c|}{FeatureIncay}&\multicolumn{4}{|c|}{Unknown}
&\multicolumn{2}{|c|}{73.37}&\multicolumn{2}{|c}{37.34}
\\
\multicolumn{4}{c|}{zy.deng1}&\multicolumn{4}{|c|}{Unknown}
&\multicolumn{2}{|c|}{72.70}&\multicolumn{2}{|c}{38.03}
\\
\multicolumn{4}{c|}{GuoGuo}&\multicolumn{4}{|c|}{Unknown}
&\multicolumn{2}{|c|}{73.55}&\multicolumn{2}{|c}{37.26}
\\
\multicolumn{4}{c|}{singlemodel}&\multicolumn{4}{|c|}{Unknown}
&\multicolumn{2}{|c|}{72.98}&\multicolumn{2}{|c}{38.24}
\\
\hline
\multicolumn{4}{c|}{DMLS (Ours)}&\multicolumn{4}{|c|}{VGG-16}
&\multicolumn{2}{|c|}{67.86}&\multicolumn{2}{|c}{28.15}
\\
\multicolumn{4}{c|}{DMLS (Ours)}&\multicolumn{4}{|c|}{ResNet-50}
&\multicolumn{2}{|c|}{\underline{75.12}}&\multicolumn{2}{|c}{\underline{40.25}}
\\
\multicolumn{4}{c|}{DMLS (Ours)}&\multicolumn{4}{|c|}{ResNet-101}
&\multicolumn{2}{|c|}{\textbf{75.34}}&\multicolumn{2}{|c}{\textbf{40.87}}
\\
\hline
\end{tabular}
}
\vspace{-4mm}
\label{table:ade20k_test}
\end{center}
\end{table}
The ADE20K dataset~\cite{zhou2017scene} is a more challenging scene parsing benchmark containing 150 semantic categories.
It comprises more than 20K natural scene images with 20,210 images for training, 2,000 images for validation and 3,352 images for testing.
The semantic categories include stuff like sky, road, grass, and discrete objects like person, car, bed.
In this dataset, images includes non-uniform distributions of objects, and the labels are inaccurate with ambiguity annotations.
Thus, it mimicks a more natural object occurrence in daily scenes.
We train our model on the training set, and evaluate it on the validation and test set.
Tab.~\ref{table:ade20k_val} shows quantitative results of compared methods on the validation set.
From the quantitative results, we can observe that the proposed models significantly outperform the FCN-based methods provided by the benchmark organizers~\cite{zhou2017scene}.
More specifically, our approach with the same VGG-16 backbone improves the DilatedNet~\cite{yu2016multi} with 7.48\% and 10.5\% in terms of pixel accuracy and mean IoU, respectively.
It outperforms the CascadeNet~\cite{zhou2017scene} with an improvement of 6.5/7.9.
Besides, our approach with the VGG-16 backbone achieves comparable results with DeepLabv2 (ResNet-101)~\cite{chen2018deeplab} and EncNet (ResNet-50)~\cite{zhang2018context}, which adopt a much deeper backbone.
This further demonstrates the effectiveness of our methods.
With the more powerful ResNet-50 backbone, our approach achieves a mean IOU score of 45.52, surpassing the outstanding PSPNet~\cite{zhao2017pyramid} and EncNet~\cite{zhang2018context}.
It also shows better performances than the very recent PSANet~\cite{zhao2018psanet}, UperNet~\cite{xiao2018unified}, GANet~\cite{zhang2019deep} and OCNet~\cite{yuan2018ocnet}.
With the ResNet-101 backbone, our method achieves best parsing results with the 83.20\% pixel accuracy and 46.16\% mean IoU.
%

To evaluate the performance on the test set, our proposed models are fine-tuned for additional 5 epochs on the train-val set.
Then we submit the predicted results of the test set to the official benchmark.
Tab.~\ref{table:ade20k_test} shows the quantitative results with previous top-ranked methods listed on the benchmark.
From the results, we can observe that our approach with the ResNet-50 backbone achieves better performance than the PSPNet~\cite{zhao2017pyramid} and EncNet~\cite{zhang2018context}.
It also shows comparable results to the ResNet-101 based GANet~\cite{zhang2019deep}.
With the ResNet-101 backbone, our approach delivers best results among all compared methods, including the unpublished models.
Visual examples on the ADE20K dataset are shown in Fig.~\ref{fig:ade20k_comparison}.
It can be observed that our parsing results contain more detailed structures compared to other baselines.
Besides, our method can relieve the semantic ambiguity problems (the second row), and parse the natural scene correctly.
\begin{table*}
\begin{center}
\doublerulesep=0.5pt
\caption{Results with different model settings on the ADE20K validation dataset. The best two results are in bold and underline.}
\resizebox{0.85\textwidth}{!}
{
\begin{tabular}{|c|c|c|c|c|c|c|c|c|c|c|c|c|c|c|c|c|c|c|c|c|c|c|c|c|c|c|c|c|||c|c|c|c|c|c|c|c|}
\hline
\multicolumn{4}{c|}{Methods}&CRF&MLS&RFCN&DSL&WCE&\multicolumn{2}{|c|}{Pixel Acc.\%}&\multicolumn{2}{|c}{Mean IoU\%}
\\
\hline
\multicolumn{4}{c|}{FCN8s~\cite{long2015fully}}&&&&&
&\multicolumn{2}{|c|}{71.32}&\multicolumn{2}{|c}{29.39}
\\
\multicolumn{4}{c|}{FCN8s~\cite{long2015fully}}&$\surd$&&&&
&\multicolumn{2}{|c|}{73.84}&\multicolumn{2}{|c}{31.45}
\\
\multicolumn{4}{c|}{FCN8s~\cite{long2015fully}}&&$\surd$&&&
&\multicolumn{2}{|c|}{75.33}&\multicolumn{2}{|c}{35.18}
\\
\hline
\multicolumn{4}{c|}{FCN8s~\cite{long2015fully}}&&&$\surd$&&
&\multicolumn{2}{|c|}{76.28}&\multicolumn{2}{|c}{36.24}
\\
\multicolumn{4}{c|}{FCN8s~\cite{long2015fully}}&&$\surd$&$\surd$&&
&\multicolumn{2}{|c|}{78.59}&\multicolumn{2}{|c}{40.62}
\\
\multicolumn{4}{c|}{FCN8s~\cite{long2015fully}}&&&$\surd$&$\surd$&
&\multicolumn{2}{|c|}{77.83}&\multicolumn{2}{|c}{38.26}
\\
\multicolumn{4}{c|}{FCN8s~\cite{long2015fully}}&&$\surd$&$\surd$&$\surd$&
&\multicolumn{2}{|c|}{79.21}&\multicolumn{2}{|c}{41.30}
\\
\hline
\multicolumn{4}{c|}{FCN8s~\cite{long2015fully}}&&&&&$\surd$
&\multicolumn{2}{|c|}{72.89}&\multicolumn{2}{|c}{31.44}
\\
\multicolumn{4}{c|}{FCN8s~\cite{long2015fully}}&&&$\surd$&&$\surd$
&\multicolumn{2}{|c|}{77.31}&\multicolumn{2}{|c}{37.52}
\\
\multicolumn{4}{c|}{FCN8s~\cite{long2015fully}}&&$\surd$&$\surd$&&$\surd$
&\multicolumn{2}{|c|}{79.51}&\multicolumn{2}{|c}{41.74}
\\
\multicolumn{4}{c|}{FCN8s~\cite{long2015fully}}&&&$\surd$&$\surd$&$\surd$
&\multicolumn{2}{|c|}{78.87}&\multicolumn{2}{|c}{39.42}
\\
\multicolumn{4}{c|}{FCN8s~\cite{long2015fully}}&&$\surd$&$\surd$&$\surd$&$\surd$
&\multicolumn{2}{|c|}{\underline{80.21}}&\multicolumn{2}{|c}{\underline{42.12}}
\\
\hline
\multicolumn{4}{c|}{DFCN (Ours)}&&$\surd$&$\surd$&$\surd$&$\surd$
&\multicolumn{2}{|c|}{\textbf{81.03}}&\multicolumn{2}{|c}{\textbf{42.85}}
\\
\hline
\end{tabular}
}
\vspace{-4mm}
\label{table:ablation_modules}
\end{center}
\end{table*}
\subsection{Ablation Studies}
To quantify the benefits of our proposed methods, we also conduct experiments with different model settings.
For the ablations, we adopt the ADE20K validation dataset, and measure the pixel accuracy and mean IoU.
For fair comparisons, we adopt the VGG-16 model~\cite{simonyan2014very} for the FCN backbone.
Thus, we could focus on reducing the variables in the empirical evaluation, and directly demonstrate the efficacy of the proposed modules.
As a starting point, we take the FCN8s~\cite{long2015fully} as the baseline, whose implementation is publicly available.
Then, we progressively add the proposed modules on top.
Note that all compared models are trained with exactly the same settings described in Subsection IV. A.

\begin{figure}
\centering
\resizebox{0.5\textwidth}{!}
{
\begin{tabular}{@{}c@{}c@{}c@{}c@{}c@{}c@{}c@{}c@{}c@{}c@{}c@{}c@{}c@{}c@{}c@{}c@{}c@{}c@{}c}
\includegraphics[width=0.25\linewidth,height=1.6cm]{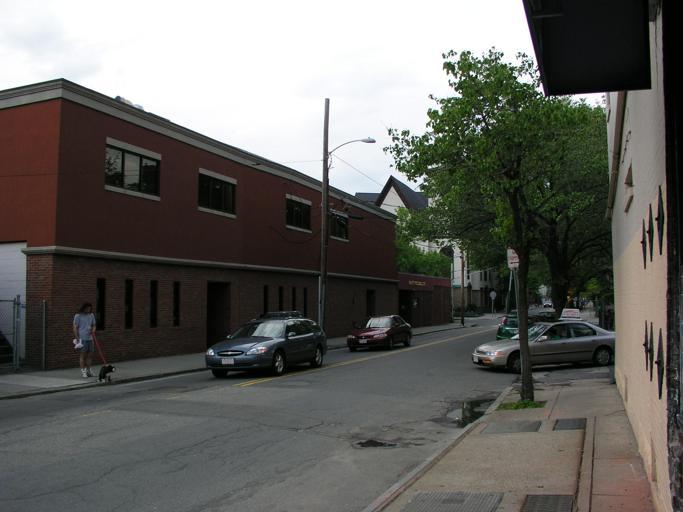}&
\includegraphics[width=0.25\linewidth,height=1.6cm]{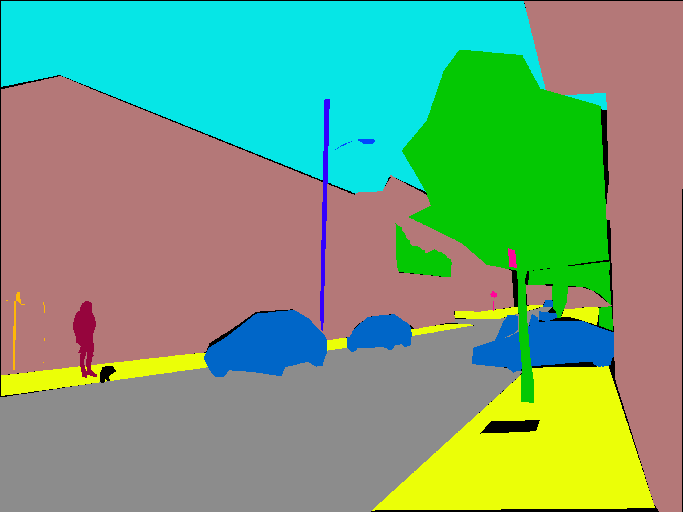} &
\includegraphics[width=0.25\linewidth,height=1.6cm]{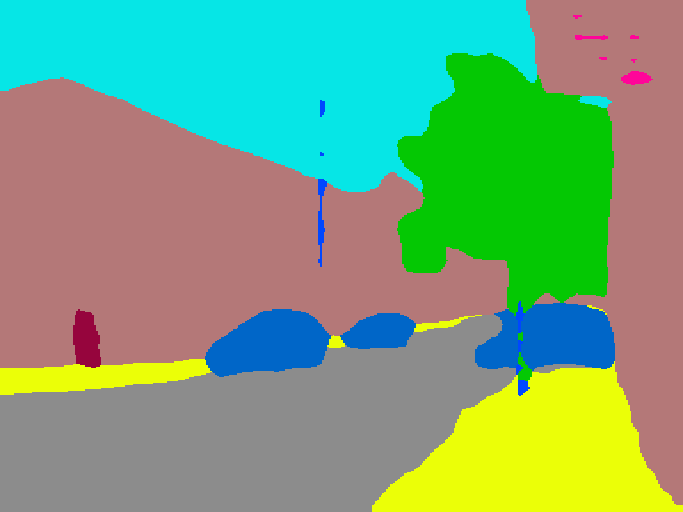} &
\includegraphics[width=0.25\linewidth,height=1.6cm]{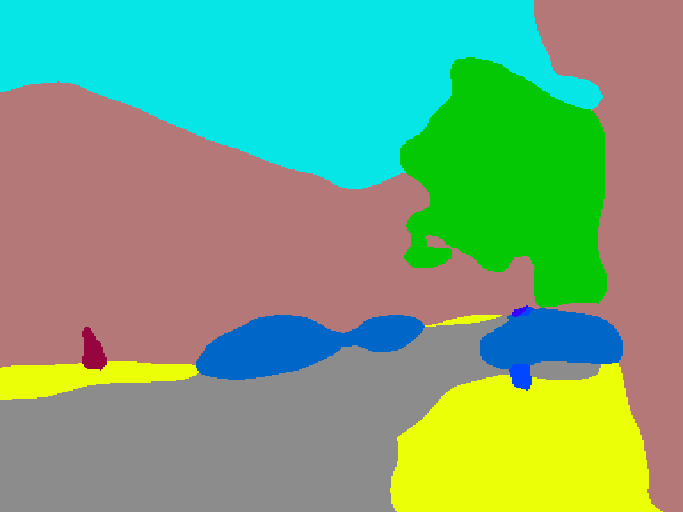}\\
{\small(a) Input} & {\small(b) Ground Truth} & {\small(c) With MLS} & {\small(d) Without MLS}\\
\end{tabular}
}
\caption{A visual example of parsing results with/without the MLS module. The prediction is based on the FCN8s~\cite{long2015fully}. More accurate boundaries can be captured with the MLS module.}
\vspace{-4mm}
\label{fig:ade20k_mls}
\end{figure}
\begin{figure*}
\centering
\resizebox{1\textwidth}{!}
{
\begin{tabular}{@{}c@{}c@{}c@{}c@{}c@{}c@{}c@{}c@{}c@{}c@{}c@{}c@{}c@{}c@{}c@{}c@{}c@{}c@{}c}
(a)\hspace{2.6cm}(b)\hspace{2.6cm}(c)\hspace{2.6cm}(d)\hspace{2.6cm}(e)\hspace{2.6cm}(f)\\
\includegraphics[width=\linewidth,height=1.45cm]{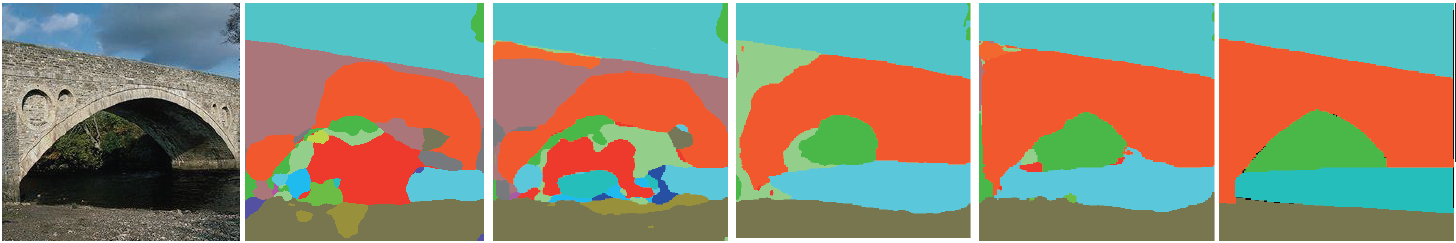} \\
\includegraphics[width=\linewidth,height=1.45cm]{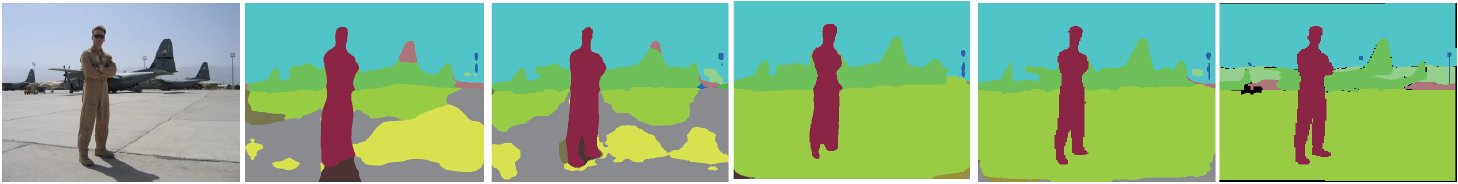} \\
\end{tabular}
}
\caption{Qualitative comparison of parsing results with different time steps. (a) Input Images. (b)-(e) Parsing results at the 1st-4rd time step, respectively. (f) Ground Truth. The quality of predictions is progressively improved. The recurrent architecture serves as a refinement mechanism to correct previous errors.}
\label{fig:ade20k_rfcn}
\end{figure*}
\begin{figure*}
\centering
\resizebox{1\textwidth}{!}
{
\begin{tabular}{@{}c@{}c@{}c@{}c@{}c@{}c@{}c@{}c@{}c@{}c@{}c@{}c@{}c@{}c@{}c@{}c@{}c@{}c@{}c}
{\small(a)} & {\small(b)} & {\small(c)} & {\small(d)} & {\small(e)} & {\small(f)}& {\small(g)}\\
\vspace{-0.5mm}
\includegraphics[width=0.15\linewidth,height=1.45cm]{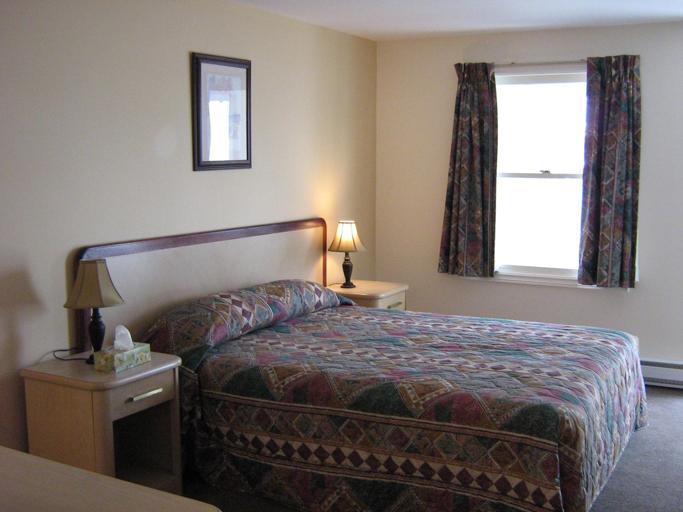}&
\includegraphics[width=0.15\linewidth,height=1.45cm]{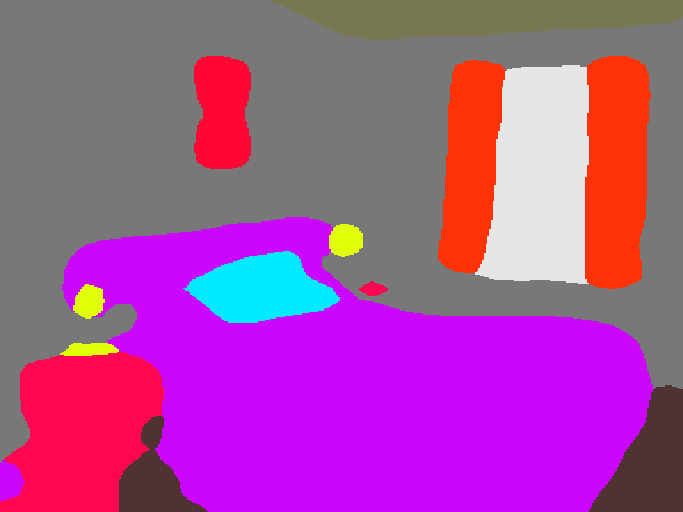} &
\includegraphics[width=0.15\linewidth,height=1.45cm]{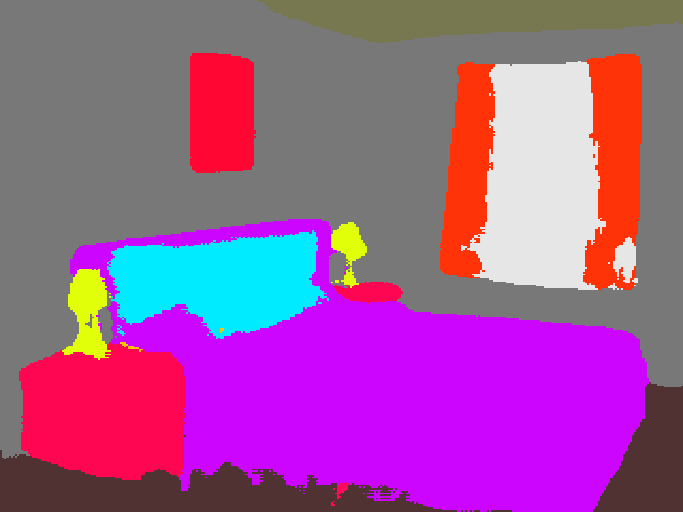} &
\includegraphics[width=0.15\linewidth,height=1.45cm]{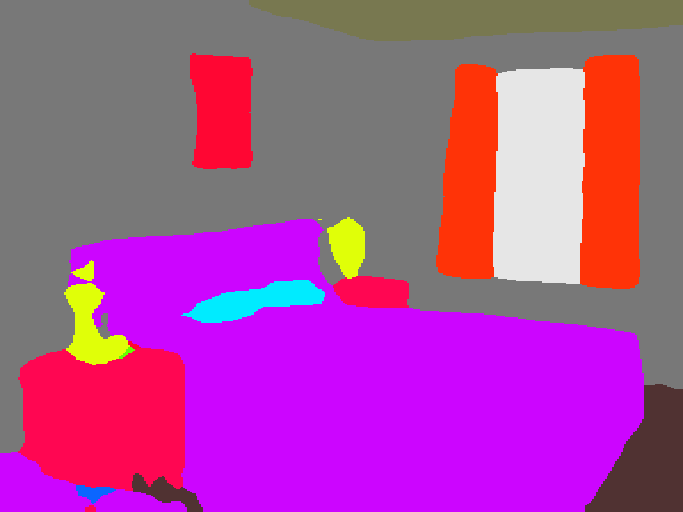} &
\includegraphics[width=0.15\linewidth,height=1.45cm]{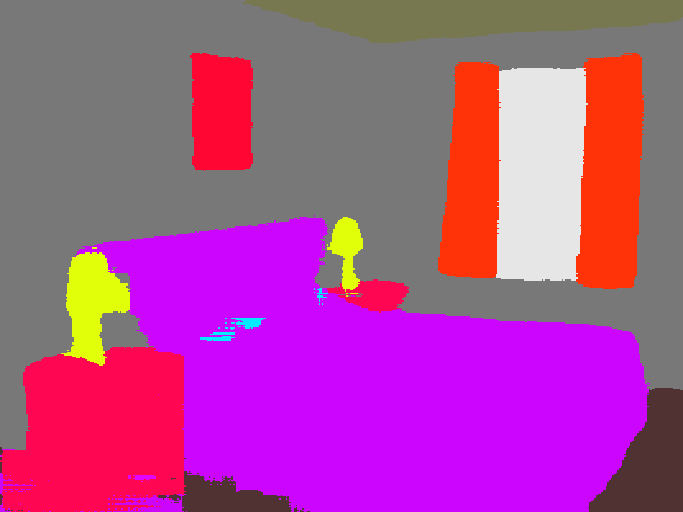} &
\includegraphics[width=0.15\linewidth,height=1.45cm]{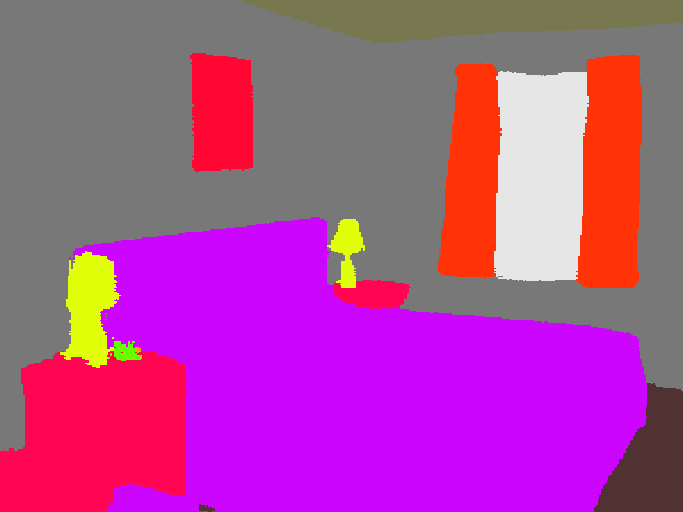} &
\includegraphics[width=0.15\linewidth,height=1.45cm]{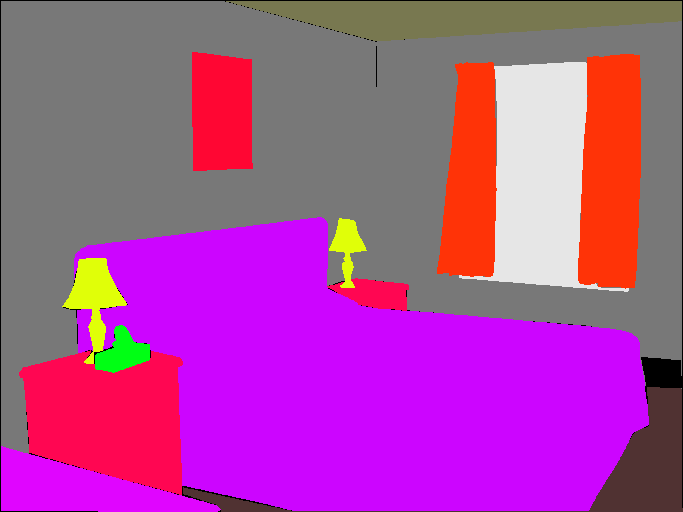} \\
\vspace{-0.5mm}
\includegraphics[width=0.15\linewidth,height=1.45cm]{imgs_ADE_val_00000034.jpg}&
\includegraphics[width=0.15\linewidth,height=1.45cm]{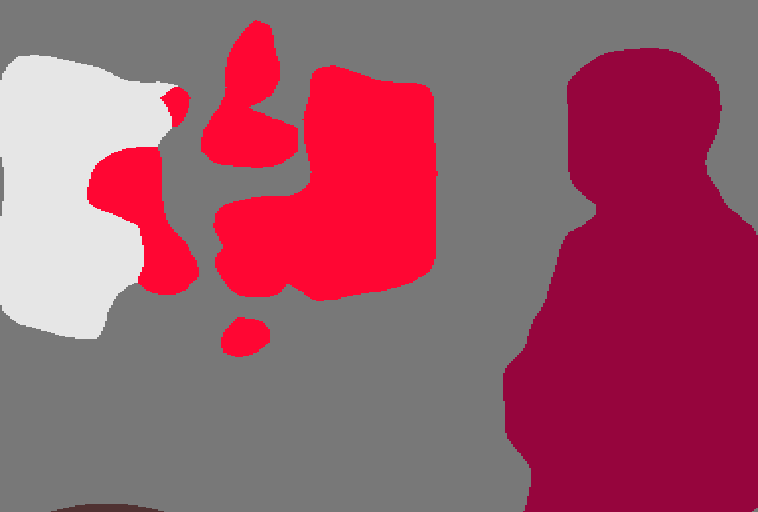} &
\includegraphics[width=0.15\linewidth,height=1.45cm]{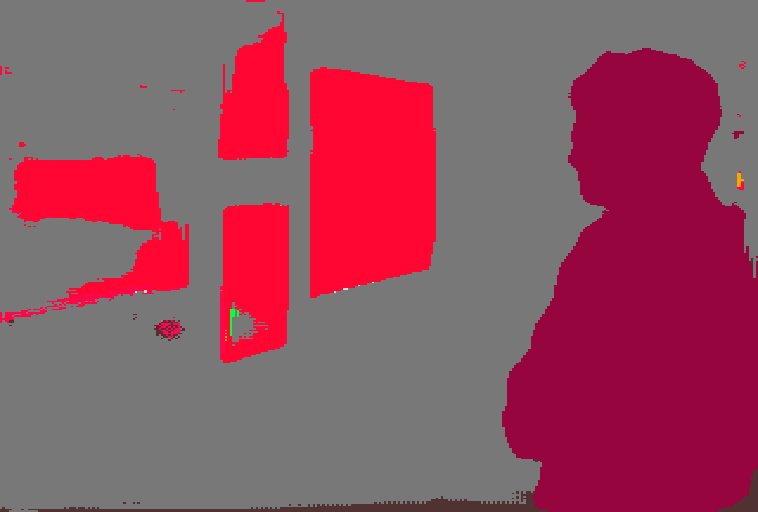} &
\includegraphics[width=0.15\linewidth,height=1.45cm]{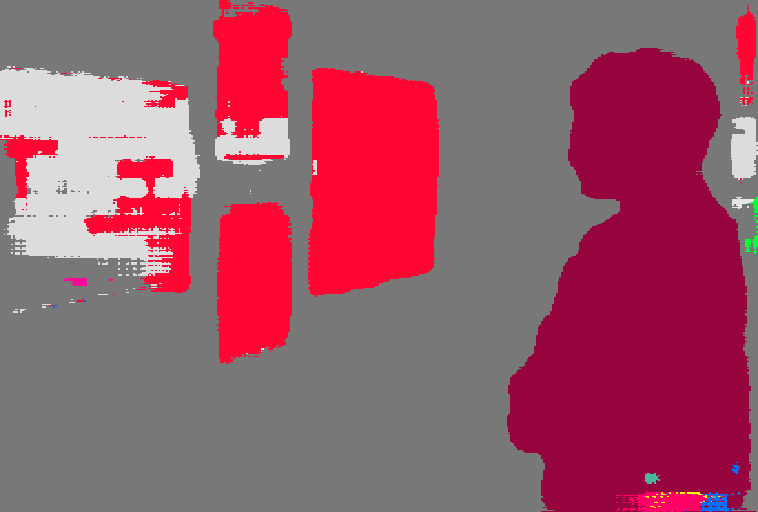} &
\includegraphics[width=0.15\linewidth,height=1.45cm]{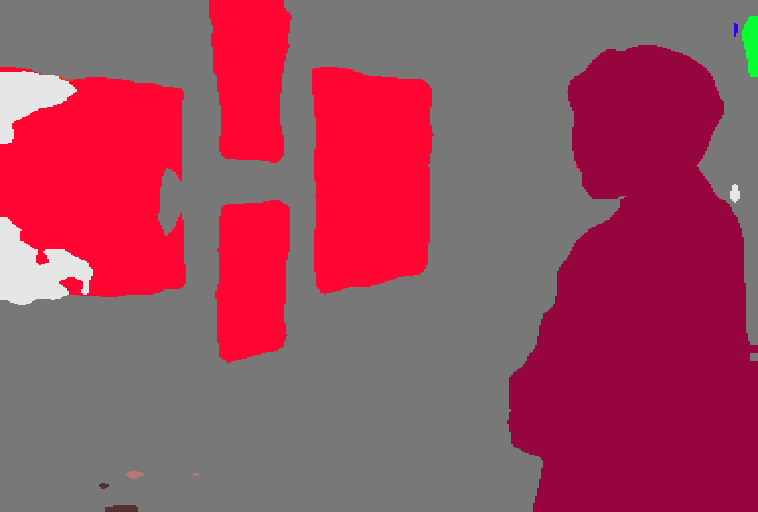} &
\includegraphics[width=0.15\linewidth,height=1.45cm]{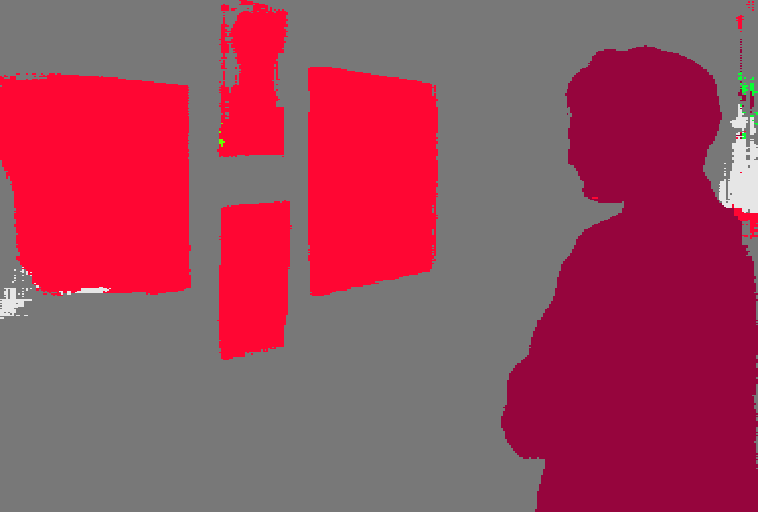} &
\includegraphics[width=0.15\linewidth,height=1.45cm]{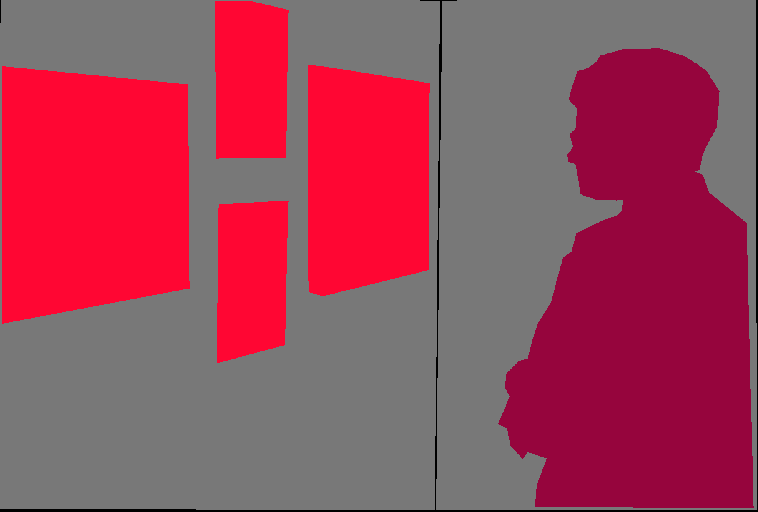} \\
\vspace{-0.5mm}
\includegraphics[width=0.15\linewidth,height=1.45cm]{imgs_ADE_val_00000113.jpg}&
\includegraphics[width=0.15\linewidth,height=1.45cm]{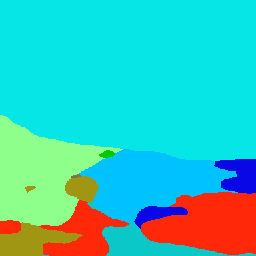} &
\includegraphics[width=0.15\linewidth,height=1.45cm]{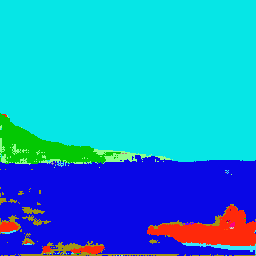} &
\includegraphics[width=0.15\linewidth,height=1.45cm]{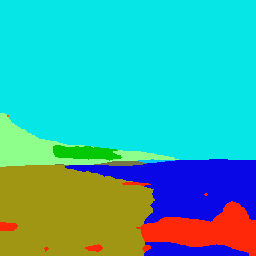} &
\includegraphics[width=0.15\linewidth,height=1.45cm]{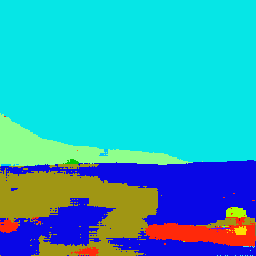} &
\includegraphics[width=0.15\linewidth,height=1.45cm]{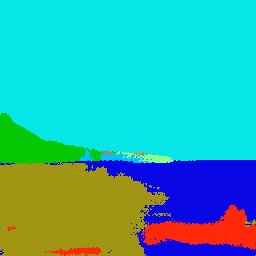} &
\includegraphics[width=0.15\linewidth,height=1.45cm]{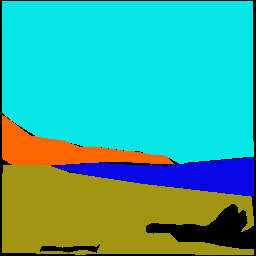} \\
\end{tabular}
}
\caption{Qualitative comparison of parsing results with/without deeply supervised learning (DSL). (a) Input Images; (b) Results with the RFCN+SCE; (c) Results with the RFCN+WCE; (d) Results with the RFCN+SCE+DSL; (e) Results with the RFCN+WCE+DSL; (f) Results with the RFCN+MLS+WCE+DSL; (g) Ground Truth.}
\vspace{-4mm}
\label{fig:ade20k_dsl}
\end{figure*}

\textbf{Effects of MLS modules.}
The proposed MLS module plays a key role in predicting contour information.
Tab.~\ref{table:ablation_modules} presents the quantitative parsing results with/without MLS modules.
Compared with the baseline FCN8s~\cite{long2015fully}, adding the proposed MLS module can yield better results of 75.33/35.18 (pixel accuracy and mean IoU), which consistently outperform the baseline results of 71.32/29.39.
The relative improvement is very expressive with about 4\%/6\% points.
It clearly demonstrates the effectiveness of the MLS module.
Besides, many existing works use the Conditional Random Fields (CRF)~\cite{krahenbuhl2011efficient} to refine the predicted results.
We also add the CRF to the FCN8s, which acts as a post-processing method.
The quantitative results are listed in the second row of Tab.~\ref{table:ablation_modules}.
Compared with the CRF, our MLS achieves better results with a relative improvement of 3.7\% in the mean IoU.
The main reason may be that the MLS implicitly captures the contour evolution, while the CRF mainly focuses on the relationships of generic pixels.
Fig.~\ref{fig:ade20k_mls} shows the visual effects of introducing the MLS.
As we can see, the MLS indeed helps the model to estimate contours, resulting more accurate parsing results.

\begin{table}
\begin{center}
\doublerulesep=0.5pt
\caption{Quantitative results with different time steps (Number of DFCNs). The best two results are in bold and underline.}
\resizebox{0.48\textwidth}{!}
{
\begin{tabular}{|c|c|c|c|c|c|c|c|c|c|c|c|c|c|c|c|c|c|c|c|c|c|c|c|c|c|c|c|c|||c|c|c|c|c|c|c|c|}
\hline
\multicolumn{4}{c|}{No. of DFCNs ($K$)}
&\multicolumn{2}{|c|}{1}&\multicolumn{2}{|c|}{2}&\multicolumn{2}{|c|}{3}&\multicolumn{2}{|c|}{4}&\multicolumn{2}{|c}{5}
\\
\hline
\multicolumn{4}{c|}{Pixel Acc.\%}
&\multicolumn{2}{|c|}{72.31}&\multicolumn{2}{|c}{76.56}&\multicolumn{2}{|c|}{77.23}&\multicolumn{2}{|c}{\textbf{77.35}}&\multicolumn{2}{|c}{\underline{77.34}}
\\
\hline
\multicolumn{4}{c}{Mean IoU\%}
&\multicolumn{2}{|c|}{31.05}&\multicolumn{2}{|c}{34.54}&\multicolumn{2}{|c|}{36.51}&\multicolumn{2}{|c}{\underline{36.84}}&\multicolumn{2}{|c}{\textbf{36.89}}
\\
\hline
\end{tabular}
}
\vspace{-4mm}
\label{table:ade20k_rfcns}
\end{center}
\end{table}
\textbf{Effects of RFCN.}
In our approach, we introduce the RFCN for sequential feature extraction.
The RFCN learns multi-level representations of input images with different spatial contexts.
To analyze the effects of RFCN, we perform two kind of experiments.
First, to evaluate the effects of the time steps, we build the RFCN structure with varied time steps.
As shown in Fig.~\ref{fig:framework}, each time step adopts a DFCN for feature extraction.
Tab.~\ref{table:ade20k_rfcns} presents the quantitative results.
As we can see, our approach only with one DFCN achieves better results than the FCN8s~\cite{long2015fully}, from 71.32/29.39 to 72.16/30.45.
This fact demonstrates the effectiveness of our DFCN architecture.
Meanwhile, the results also show that introducing recurrent connections boosts the parsing performance.
The performance is steadily improved at the first three steps.
However, we observe that the performance almost converges after the 4-th time step.
Besides, adding the time steps increases the model size and computation.
Therefore, we set the total time step of the RFCN to $K = 4$.
Fig.~\ref{fig:ade20k_rfcn} shows the visual results with different time steps.
It can be observed that the multiple recurrent architectures mainly serves as a refinement mechanism to correct previous errors.
The coarse results contain rich context information, possibly helping more accurate predictions.

In addition, we also study the joint effects of the RFCN and MLS modules.
In this setting, we adopt the RFCN with four time steps.
Tab.~\ref{table:ablation_modules} shows the quantitative results.
The proposed RFCN achieves much better performance than the baseline FCN8s~\cite{long2015fully}, from 71.32/29.39 to 76.28/36.24.
This fact further proves the effectiveness of recurrent structures.
Combined with the MLS modules, the performance can be improved with a considerable margin (2.3/4.2).
The resulting method achieves the scores of 78.59/40.62 in terms of pixel accuracy and mean IoU.

\textbf{Effects of DSL.}
In our approach, we adopt the deeply supervised learning to guide the network updates.
For the loss function, we utilize the weighted cross-entropy (WCE), while most of existing methods adopt the standard softmax cross-entropy (SCE) loss.
To verify the effects of the DSL, we first train our network model with standard SCE losses.
It makes our work more directly comparable with the published approaches.
The quantitative results are also shown in Tab.~\ref{table:ablation_modules}.
Note that the baselines in the 1-5 rows adopt a single SCE loss for the model optimization.
By introducing the DSL, our RFCN (the 6-th row) can achieve scores of 77.83/38.26, which are competing to the DeepLabv2~\cite{chen2018deeplab}, MOE-SPNet~\cite{fu2018moe} and GANet~\cite{zhang2019deep}.
It outperforms the single-loss based RFCN (the 4-th row) with an improvement of 1.5/2.0.
When combined with MLS modules, our approach shows much better results with an improvement of 1.4/3.0, resulting to 79.21/41.34.
The results are in par with the EncNet~\cite{zhang2018context} and UperNet~\cite{xiao2018unified}.

To further verify the effects of the DSL, we replace the standard SCE loss with the proposed WCE loss, and keep other components the same.
The 8-12 rows in Tab.~\ref{table:ablation_modules} show the quantitative results.
Compared with the above results, the models with the WCE loss can constantly achieve better parsing performance with a considerable margin.
We also observe the same performance trend as appeared in the SCE setting.
When including all proposed modules, the complete model yields best results (81.03/42.85), exceeding the method without DSL (78.59/40.62) by a large margin (2.5/2.2).
Thus, for scene parsing it is benefit with appropriate loss functions and learning methods for performance improvements.
Fig.~\ref{fig:ade20k_dsl} shows the visual results with/witout DSL. From the results, we can see that the quality of predictions is improved with the WCE and DSL.
\begin{table*}
\begin{center}
\caption{Run-time analysis of typical methods. It is conducted with an i4790 CPU and an NVIDIA Titan X.}
\label{table:times}
\resizebox{1\textwidth}{!}
{
\begin{tabular}{|c|c|c|c|c|c|c|c|c|c|c|c|c|c|c|c|c|c|c|c|c|c|c|c|}
\hline
Models
&Ours
&SegNet~\cite{badrinarayanan2017segnet}
&FCN8s~\cite{long2015fully}
&DeepLabv2~\cite{chen2017rethinking}
&DilatedNet~\cite{yu2016multi}
&RefineNet~\cite{lin2016efficient}
&PSPNet~\cite{zhao2017pyramid}
&DenseASPP~\cite{yang2018denseaspp}
&DeepLabv3+~\cite{chen2018encoder}
&GANet~\cite{zhang2019deep}
\\
\hline
Speed (fps)
&6.53
&12.87
&16.24
&3.37
&9.21
&0.07
&2.89
&1.77
&2.61
&1.78
\\
\hline
Flops (G)
&102.4
&286.0
&220.4
&578.1
&382.6
&2832
&722.1
&498.4
&748.2
&216.8
\\
\hline
Parameters (M)
&34.0
&29.5
&134.5
&44.0
&140.8
&134.0
&128.7
&15.2
&26.8
&148.5
\\
\hline
\end{tabular}
}
\vspace{-6mm}
\end{center}
\end{table*}
\subsection{More Discussion}
\textbf{Run-time analysis.}
Speed of an algorithm is an important factor for practical applications.
We conduct additional experiments on run-time analysis of different methods.
Note that different methods adopts different devices and toolboxes, and it is hard to measure all compared methods directly.
We adopts the source codes of compared methods, and evaluate them on an NVIDIA Titan X GPU for this measure.
For a fair comparison, we choose the Cityscapes dataset and adopt the 512$\times$512 as the resolution of the input image.
In this process, we don't employ any testing technology, like multi-scale or multi-crop testing.
Tab.~\ref{table:times} shows the statistics of speed, flops and parameters.
The flops and parameters indicate the number of operations to process images.
From the results, we can observe that our method achieves significant progress against several methods both in speed and computational complexity.
For example, our method runs fast than strong baseline methods such as PSPNet, DenseASPP, DeepLabv3+.

\textbf{Extensions with more powerful backbones.}
Many existing works have shown that deeper neural networks deliver better performance to large-scale image classification and segmentation tasks.
In our experiments, we also observe this trend on the three public benchmarks.
With more powerful backbones (from VGG-16 to ResNet-50 and ResNet-101), our model can consistently achieve better performance, as shown in Tab. 1-4.
To further improve the performance, one can extend our model with very deep backbones, \emph{e.g.}, ResNet-152, ResNeXt-101/152, DenseNet-121/161.
However, this inevitably requires more computation burden and inference time, which might harm them for practical applications.
Meanwhile, our approach with relatively shallow backbones can show comparable even better performance than methods that use deeper backbones, like PSPNet and DeepLabv2.
Our DFCN plays a considerable role in achieving this target, as shown in Tab.~\ref{table:ade20k_rfcns} and  Tab.~\ref{table:ablation_modules}.
Thus, building more powerful yet lightweight backbones is also a valuable direction for model extensions.
%

\textbf{Important information of complex scenes.}
Semantic scene parsing outputs dense labeling maps for natural images.
As demonstrated in our experiments, the proposed approach is able to correctly parse semantic categories which occupy large scene areas.
%
%
This partly proves the feasibility of our proposed level set methods.
For practical applications, we may focus on some important objects, such as pedestrians, cars, street light, bicyclist and sign poles.
However, our approach is still not perfect in these categories, as shown in Tab.~\ref{table:cityscapes_test_cls}.
Thus, incorporating the importance information to these categories may improve the parsing performance.
For complex scenes, possible solutions are designing more powerful structural output models, or training the models with related tasks, such as object detection and image caption.
\section{Conclusion and Future Work}
In this paper, we propose a novel end-to-end learning framework, named Deep Multiphase Level Set, for semantic scene parsing.
It consists of three components, \emph{i.e.}, recurrent FCNs, adaptive multiphase level set, and deeply supervised learning.
Instead of learning a level set function for binary classifications, the framework utilizes multiphase level sets to model spatial dependencies and boundary refinements for multi-class scene parsing.
The proposed framework is continuously differentiable, thus it can be applied to any fully convolutional networks.
Furthermore, recurrent FCNs are proposed to extract multi-level features sequentially, which encode rich spatial and contextual information.
To enhance the results, we adopt deeply supervised learning to effectively train the models.
Extensive experiments demonstrate that our approach is a cutting-edge solution, achieving the scene parsing effectively and efficiently.
As a challenging task, scene parsing is still faced with many problems, such as needing large-scale dense annotations.
In the future, we will exploit the weak annotations or unsupervised methods to handle the scene parsing task.
\ifCLASSOPTIONcaptionsoff
  \newpage
\fi
\bibliographystyle{IEEEtran}
\bibliography{IEEEabrv,refs}
\end{document}